\documentclass[lettersize,journal]{IEEEtran}
\usepackage{amsmath,amsfonts}
\usepackage{algorithmic}
\usepackage{algorithm}
\usepackage{array}
\usepackage[caption=false,font=normalsize,labelfont=sf,textfont=sf]{subfig}
\usepackage{textcomp}
\usepackage{stfloats}
\usepackage{url}
\usepackage{verbatim}
\usepackage{graphicx}
\usepackage{cite}
\usepackage{caption}
\usepackage{subcaption}
\usepackage{multirow}
\usepackage[table]{xcolor}
\usepackage{wrapfig}
\usepackage{subfig}
\usepackage{makecell}
\usepackage[export]{adjustbox}
\usepackage{booktabs}
\usepackage{bm}
\usepackage{pdfrender}
\usepackage{hyperref}

\begin{document}

\title{Magnifying change: Rapid burn scar mapping with multi-resolution, multi-source satellite imagery}

\author{Maria Sdraka, Dimitrios Michail, Ioannis Papoutsis
\thanks{M. Sdraka and I. Papoutsis are with Orion Lab, National Observatory of Athens \& National Technical University of Athens, Greece.}
\thanks{M. Sdraka and D. Michail are with the Department of Informatics and Telematics, Harokopio University of Athens, Greece.}
}

\maketitle

\begin{abstract}
Delineating wildfire affected areas using satellite imagery remains challenging due to irregular and spatially heterogeneous spectral changes across the electromagnetic spectrum. While recent deep learning approaches achieve high accuracy when high-resolution multispectral data are available, their applicability in operational settings, where a quick delineation of the burn scar shortly after a wildfire incident is required, is limited by the trade-off between spatial resolution and temporal revisit frequency of current satellite systems.
To address this limitation, we propose a novel deep learning model, namely BAM-MRCD, which employs multi-resolution, multi-source satellite imagery (MODIS and Sentinel-2) for the timely production of detailed burnt area maps with high spatial and temporal resolution. Our model manages to detect even small scale wildfires with high accuracy, surpassing similar change detection models as well as solid baselines. All data and code are available in the GitHub repository: \url{https://github.com/Orion-AI-Lab/BAM-MRCD}. 
\end{abstract}


\section{Introduction}
\label{sec:intro}

\IEEEPARstart{W}{ildfires} have become an increasingly global phenomenon, with devastating and long-lasting effects on ecosystems and communities. Human-induced climate change increases the frequency and intensity of weather conditions that facilitate the presence of wildfires --- combinations of high temperatures, low humidity and precipitation, and often high winds --- resulting in longer fire seasons and a higher risk of extreme wildfire events \cite{cunningham2024increasing, jones2020climate}. For example, heat-induced fire weather types in the Mediterranean Basin are projected to escalate by the end of the century \cite{ruffault2020increased, essa2023drought, zittis2022climate, lionello2018relation, jones2022global}, posing an imminent threat to already heavily impacted regions. Located in the northeast region of the Mediterranean basin, Greece has been suffering from devastating wildfires over the past two decades  \cite{koutsias2012relationships, papavasileiou2022catastrophic} and studies show that the risk of wildfire in the Greek terrain is bound to increase in the future \cite{rovithakis2022future, malisovas2023assessing}. Therefore, a timely assessment of the damage inflicted by wildfires is of utmost importance for the immediate planning of response and recovery efforts, in order to secure the effective restoration of the ecosystem.

\begin{figure}[h]
    \centering
    \includegraphics[width=9.8cm]{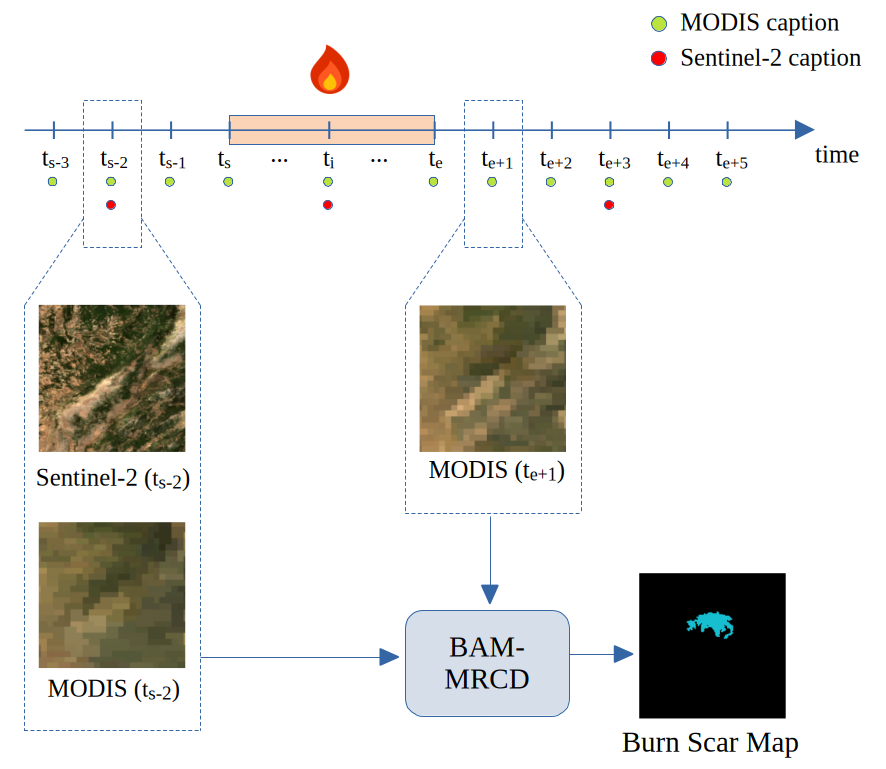}
    \caption{Outline of our approach. A wildfire event starts at timestep $t_s$ and ends at $t_e$. Due to the inherent limitations of the satellite missions, we have daily MODIS captions over the area and sparser Sentinel-2 captions. The proposed method takes advantage of the first available MODIS imagery after the fire ends ($t_{e+1}$), as well as a pair of MODIS and Sentinel-2 captions on the same date, anytime before the fire starts (e.g. $t_{s-2}$), and produces a high resolution delineation of the burn scar. This approach alleviates the need for bitemporal high resolution imagery, thus obtaining timely and accurate burn scar maps soon after a wildfire is deemed complete.}
    \label{fig:outline}
\end{figure}

Traditional ground-based mapping techniques are often limited by accessibility, time constraints, and resource availability. Satellite remote sensing has emerged as a critical tool for post-wildfire assessment, offering a means to quickly and efficiently delineate burnt areas across vast and often inaccessible regions. However, satellite sensors are characterized by distinct spectral, spatial and temporal resolutions, whereas technical limitations usually impose a trade-off between these three dimensions. In an operational setting where prompt assessment of the inflicted damage is a hard requirement for immediate relief efforts, a wildfire monitoring system should be able to leverage the advantages offered by different satellite missions to produce a delineation of the burn scar with the highest level of detail as soon as possible.

Among the various satellite sensors available for public use, the Moderate Resolution Imaging Spectroradiometer (MODIS) by NASA and Sentinel-2 by ESA demonstrate a strong potential for this task, each with their own strengths and limitations. MODIS provides frequent observations at moderate spatial resolution, allowing near real-time monitoring of fire impact, while Sentinel-2 offers high spatial resolution imagery that enhances the accuracy of the burn scar perimeter delineation, but with a lower revisit frequency spanning over a few days. This study presents a novel technique which integrates data from these two sensors to improve the timeliness and precision of burnt area mapping, employing the FLOGA dataset \cite{sdraka2024floga} which provides high-resolution manual annotations for various wildfire events in Greece accompanied by multi-source bitemporal satellite imagery. The proposed approach leverages the complementary strengths of the Sentinel-2 and MODIS satellites, i.e. the high temporal resolution of MODIS and the superior spatial resolution of Sentinel-2, to generate accurate burn scar maps shortly after a wildfire (Fig. \ref{fig:outline}). By utilizing a custom deep learning architecture, our method aims to provide an efficient and reliable solution for timely wildfire impact assessment, capable of handling burn scars of varying size. We hope that our work will propel further research in this newly emerged field and contribute to the next-day planning of relief and ecosystem restoration efforts. In short, our contributions are summarized below:

\begin{itemize}
    \item We formulate a Burnt Area Mapping (BAM) problem with multi-resolution satellite input for change detection, which is distinct from generic change detection tasks.
    \item We propose a new deep learning approach, namely BAM-MRCD, which exploits multi-scale feature extraction from pre-fire Sentinel-2 and bitemporal MODIS imagery. Our model processes both input sources in a parallel fashion, without the need to use super-resolution modules which may potentially impute noise to the pipeline.
    \item BAM-MRCD outperforms the baselines as well as other models for multi-resolution change detection proposed in the literature, achieving high accuracy on wildfire impacted areas of varying size.
\end{itemize}

\section{Related work}
\label{sec:related}

\subsection{Change detection with multi-resolution input}
\label{subsec:cd-sr}

Change detection (CD) tasks typically require a bitemporal input which comprises one image before and one image after the change under investigation. Time restrictions surfacing in operational emergency settings, such as during or shortly after natural disasters, often dictate the use of different satellite sources in order to obtain an initial assessment of the damage as soon as possible. However, this can lead to pairs of images with vastly different spatial and/or spectral resolution, thus hindering the accuracy of traditional change detection algorithms that are based on similar input pairs. To tackle this issue, various change detection approaches utilising multi-resolution satellite imagery have been proposed in the literature over the past years. A number of these studies employ a two-step approach, combining a super-resolution (SR) module for the enhancement of the low-resolution image \cite{sdraka2022deep}, along with a separate change detection module for the computation of the final output. For example, RACDNet \cite{tian_racdnet_2022} comprises a WDSR model \cite{yu2018wide} enhanced with corner detection which is initially trained for the SR task, and a siamese UNet \cite{ronneberger2015u} trained for the CD task. In a similar fashion, MF-SRCDNet \cite{li_mf-srcdnet_2023} employs a Res-UNet \cite{cao2020improved} to downscale the low resolution imagery and a modified STANet \cite{chen2020spatial} for the production of the binary change map which takes as input an auxiliary layer with edge information. Other studies propose an end-to-end approach where both SR and CD modules are trained simultaneously through a carefully designed compound loss, such as SRCDNet \cite{liu_super-resolution-based_2022} and ESR-DMNet \cite{li_esr-dmnet_2024}. Finally, a hybrid approach is followed by DCILNet \cite{li_dual-branch_2025}, where an SR module is used for resolution alignment and the output is fed to a dual-branch CD module with custom attention mechanisms fusing the extracted features of each level for facilitating the detection of multi-scale changes.

However, these approaches often suffer from the propagation of errors stemming from the SR process which is an inherently difficult task with a one-to-many mapping \cite{sdraka2022deep}. When the downscaling of the low resolution image fails or produces a significant number of artifacts, the CD process is heavily affected and false alarms are favoured. Furthermore, enhancing the resolution of every possible structure in a remote sensing image is often unnecessary since the change usually happens in a small portion of the depicted scene. For that reason, a number of end-to-end approaches have been proposed which circumvent the use of a dedicated SR module. For example, SUNet \cite{shao_sunet_2021} employs a UNet with detected edges as auxiliary input, and MM-Trans \cite{liu2022learning} comprises a cascade of transformer architectures for the spatial alignment and semantic feature extraction of both input images. Additionally, HiCD \cite{pang_hicd_2024} employs a knowledge distillation strategy from a teacher model trained exclusively on high-resolution input and the authors also propose a novel loss function which comprises various factors measuring intra- and inter-image correlation. Finally, in SILI \cite{chen_continuous_2023} a custom data synthesis pipeline is proposed with random up/downsamplings of both input images and a random swapping of image regions. In addition, extracted edges and the relative location of low-resolution pixels with respect to high-resolution pixels are fed as auxiliary information in an MLP which produces the final map.

An unsupervised method has also been proposed for the task of change detection with multi-resolution input. In \cite{zheng_unsupervised_2022}, two clustering algorithms are first used to produce clusters of adjacent pixels that are spectrally similar, called superpixels. Subsequently, each image is segmented individually and cross-resolution difference maps are produced and combined to a final binary change map.

\subsection{Burn scar mapping with multi-resolution input}
\label{subsec:bam-srcd}

The most common application of change detection methods is related to changes in buildings and infrastructure, where additional geometric information such as edges, corners and contours usually provide valuable insights to the models. Such strict geometric structure is absent in wildfires, which typically result in loosely defined burnt regions with spaces of varying burn severity. Furthermore, burn scars have a stronger spectral signature in the near-infrared and shortwave infrared ranges of the e/m spectrum, thus techniques based solely on the visible spectral bands --- which is the case in generic change detection models --- are not directly applicable. Therefore, dedicated burn scar mapping algorithms have been recently proposed which take advantage of multi-resolution satellite imagery for the timely delineation of the affected areas. For example, SSIASC \cite{du2025burned} utilises super-pixel segmentation techniques and Particle Swarm Optimization \cite{488968} on post-fire imagery in order to produce burn scar maps for areas with limited data, however at the cost of computational efficiency and thorough hyperparameter tuning. On the contrary, DLSR-FireCNet \cite{seydi_dlsr-firecnet_2025} is a deep learning model which takes as input MODIS bitemporal imagery and produces burn scar maps in Landsat spatial resolution. The model is an HR-Net \cite{wang2020deep} extended with a cascade siamese structure and trained on a loss function that allows custom penalization of false negative and false positive predictions. However, DLSR-FireCNet was solely trained and evaluated on big wildfire events ($>200$ ha) each corresponding to multiple pixels of a MODIS image, which facilitates the detection task and does not explore the applicability of the algorithm in real-case, operational settings where the accurate delineation of wildfires of any size is a hard requirement.
Finally, a recent study \cite{nolde2025multi} proposed a complex pipeline for near real-time monitoring of wildfires using varying resolution satellite imagery, along with hotspot and land cover information. The pipeline comprises hotspot clustering, super-pixel segmentation, an ensemble of Graph Convolutional Neural Networks and various filters for the exclusion of false alarms. Nevertheless, this approach requires heavy hyperparameter tuning in each step, and the end result is significantly affected by cloud and topography shadows, cirrus clouds and land cover type. In addition, the medium-resolution hotspot data used in the study hinder performance when no hotspots can be detected due to cloud coverage or when combined with higher-resolution satellite imagery for the burn scar extraction.

\section{Problem formulation}
\label{sec:problem}

Generic change detection approaches proposed in the literature are mostly focused on artificial structures, such as buildings and infrastructure, with clearly defined outlines, and easily detected in the visible spectra. Burn scars however manifest in different ranges of the e/m spectrum, with the strongest signal in the red, near-infrared and shortwave infrared bands, and their borders are often not as easily distinguishable from neighboring unburnt land. In addition, the variable severity of the wildfire further complicates the delineation of the impacted area, since the change to be detected is not homogeneous across the whole region \cite{nolde2025multi}. Fire severity is an established term referring to the degree of damage caused by the wildfire and is a product of the fire intensity, residence time and land cover type \cite{han2021quantifying}. Thus, the region inside a wildfire perimeter usually comprises multiple sub-regions with varying degrees of damage: from completely unburnt vegetation or slightly scorched understory to completely burnt flora and eroded soils. Finally, wildfire affected land often displays a spectral signature similar to crop harvesting \cite{hall2016modis} resulting in a significant number of false positive predictions \cite{sdraka2024floga}.

Therefore, burn scar mapping is an application that presents additional challenges, distinct from typical change detection problems found in the literature. In this work we investigate the use of multi-resolution satellite input imagery with the aim to provide a rapid delineation of the inflicted damage and assist first response efforts. Regarding the spatiotemporal resolution of publicly available satellite sources, we assume the presence of low- and high-resolution imagery at different timesteps $t_{i}$, both before ($t_{1}$) and after ($t_{2}$) a wildfire event. Let us denote high-spatial low-temporal resolution (HSLT) imagery captured on time $t_{1}$ as $I^{HSLT}_{t_1} \in \Re^{C_{1} \times H \times W}$, and low-spatial and high-temporal resolution (LSHT) satellite imagery captured on time $t_{1}$ as $I^{LSHT}_{t_1} \in \Re^{C_{2} \times \frac{H}{s} \times \frac{W}{s}}$. $H$ is the height of the image, $W$ the width of the image, and $s$ is the scaling factor between the two images. $C_{1}$ and $C_{2}$ is the number of channels of the HSLT and the LSHT images respectively. We also assume a learning target $p \in \mathbb{Z}_{2}^{H \times W}$, where $\mathbb{Z}_{2}$ is the binary set $\{0, 1\}$ (depicting the burnt pixels as 1 and the unburnt pixels as 0). Our goal is to define a model $M_\theta$ with $\theta$ trainable parameters which will produce a binary map $\hat{p} \in \mathbb{Z}_{2}^{H \times W}$ as close to the ground truth label $p$ as possible. To achieve a next-day burn scar mapping, we also impose the constraint that the post-image input must have a low spatial and high temporal resolution, i.e. $I_{t_{2}}^{LSHT}$. 

Therefore, we define two distinct change detection tasks based on the type of pre-fire input imagery: (i) when an LSHT pre-fire image is used (Eq. \ref{eq:lrcd}), (ii) when an HSLT pre-fire image is used (Eq. \ref{eq:mrcd}).

\begin{equation}
    Approach ~1: ~ M_{\theta}(I^{LSHT}_{t_1}, ~ I^{LSHT}_{t_2}) = \hat{p}
\label{eq:lrcd}
\end{equation}

\begin{equation}
    Approach ~ 2: ~ M_{\theta}(I^{HSLT}_{t_1}, ~ I^{LSHT}_{t_2}) = \hat{p}
\label{eq:mrcd}
\end{equation}

Due to a possibly large scaling factor between the two satellites and the inherent differences in the operational bands, information between $I^{HSLT}_{t_1}$ and $I^{LSHT}_{t_2}$ may not be directly comparable. Therefore, we opt to keep the $I^{HSLT}_{t_1}$ image, which can potentially provide high frequency details of the underlying scene, and also include a pre-fire LSHT image on time $t_{1}$, $I^{LSHT}_{t_1}$, in order to allow the model to derive additional information from contrasting the bitemporal imagery captured by this sensor. As a hard requirement, the pre-fire images should be acquired on the same day, and as close in time as possible. More formally, an additional goal is defined:
\begin{equation}
    Approach ~ 3: ~ M_{\theta}(I^{HSLT}_{t_1}, ~ I^{LSHT}_{t_1}, ~ I^{LSHT}_{t_2}) = \hat{p}
    \label{eq:mrcd_3inp}
\end{equation}

In this study, we present a novel deep learning method for Approach 3, which manages to overcome the limitations imposed by the different sensors and achieve a high quality mapping of the observed burn scar. To our knowledge, this is the first time that such an approach is proposed. We also benchmark our method against baseline and state-of-the-art models developed for Approaches 1 and 2, and discuss the obtained results.

\begin{figure}[t]
    \centering
    \includegraphics[width=0.48\textwidth]{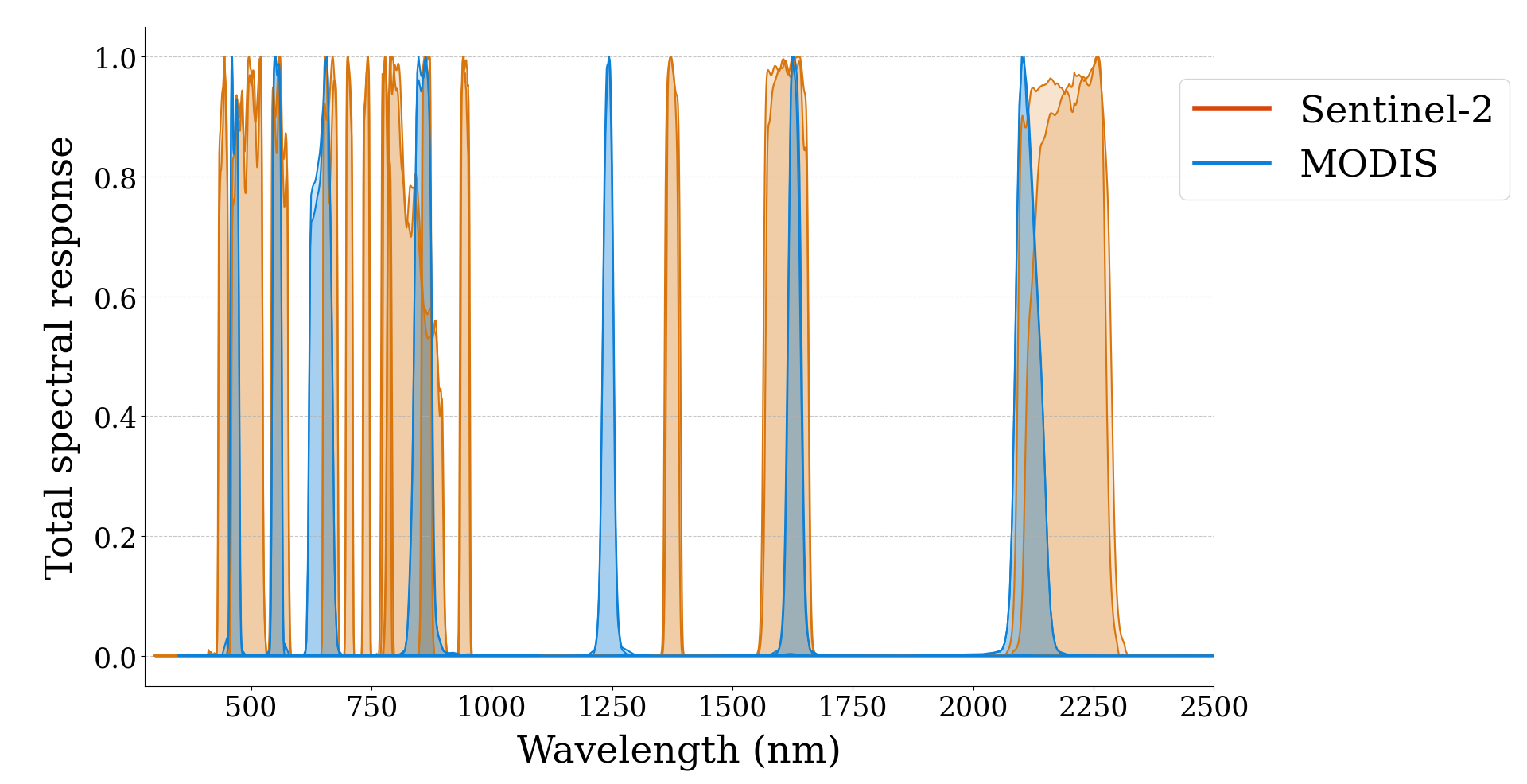}
    \caption{Normalized spectral responses of the Sentinel-2 and MODIS satellites.}
    \label{fig:spectral_responses}
\end{figure}

\begin{table}[t]
\centering
\caption{Common bands between Sentinel-2 and MODIS (the central wavelengths are given inside parentheses (in nm)).}
    \label{tab:common_bands}
\begin{tabular}{m{1.2cm}m{2.5cm}m{2.5cm}}
    \toprule
      Band & Sentinel-2 (A/B) & MODIS (Terra/Aqua) \\
    \midrule
    Blue & B02 (492.4/492.1) & B03 (469) \\
    Green & B03 (559.8/559.0) & B04 (555) \\
      Red & B04 (664.6/664.9) & B01 (645) \\
    NIR & B08 (832.8/832.9) & B02 (858.5) \\
    NIR & B8A (864.7/864.0) & B02 (858.5) \\
    SWIR & B12 (2202.4/2185.7) & B07 (2130) \\
    \bottomrule
\end{tabular}
\end{table}

\section{Data}
\label{sec:data}

For the purpose of this study we utilise FLOGA, an analysis-ready dataset for burn scar mapping which contains bitemporal satellite imagery for 326 wildfire events in Greece over a 5-year time span (2017-2021). The total area of each event ranges from $\sim$22 $ha$ (0.22 $km^{2}$) to $\sim$45,284 $ha$ (452.84 $km^{2}$). In particular, FLOGA provides a pair of pre- and post-fire Sentinel-2 L2A captions, along with MODIS surface reflectance imagery (MOD09GA) on the same dates. Sentinel-2 (A/B) images have 13 channels at 10m, 20m, and 60m Ground Sampling Distance (GSD), while MODIS (Terra/Aqua) images have 7 channels at 500m GSD. The spectral responses of both satellite constellations can be seen in Fig. \ref{fig:spectral_responses}. All images have been selected in such a way as to minimize the presence of clouds, haze and/or smoke over the burnt areas, and have been aligned to a common spatial grid by nearest neighbor interpolation.

The labels in FLOGA have been produced through careful photointerpretation by a group of experts in the Hellenic Fire Service based primarily on Sentinel-2 observations at 10m GSD. Obtaining annotations from the end users of a potential automatic burn scar mapping system renders FLOGA an essential tool for the development of robust and accurate algorithms capable of being adopted at an operational level.

To facilitate model design and convergence, we interpolate all Sentinel-2 bands to 60m GSD, thus settling on a $\sim8\times$ scaling factor between the Sentinel-2 and MODIS images ($s = 8$).
Subsequently, we crop the images into non-overlapping patches of size 256$\times$256 pixels and split the patches into train, validation and test sets. To establish the absence of information leak between the different sets, we opt for a spatial split based on the wildfire events. After filtering patches based on the bitemporal MODIS image quality (absence of artifacts, clouds, etc) and imposing the requirement that the total affected area must be larger than a single MODIS pixel (i.e. $> 25$ ha), we keep 173 events for training, 85 for validation and 53 for testing. Since natural disasters are natively rare phenomena, positive samples (i.e. containing at least one burnt pixel) are much limited in number than the non-positive, resulting in a highly imbalanced setting. Therefore, we randomly sample one negative patch for each positive included in each set, resulting in a total of 454 training, 208 validation and 132 test patches.

\begin{figure*}[h]
    \centering
    \includegraphics[width=12.6cm]{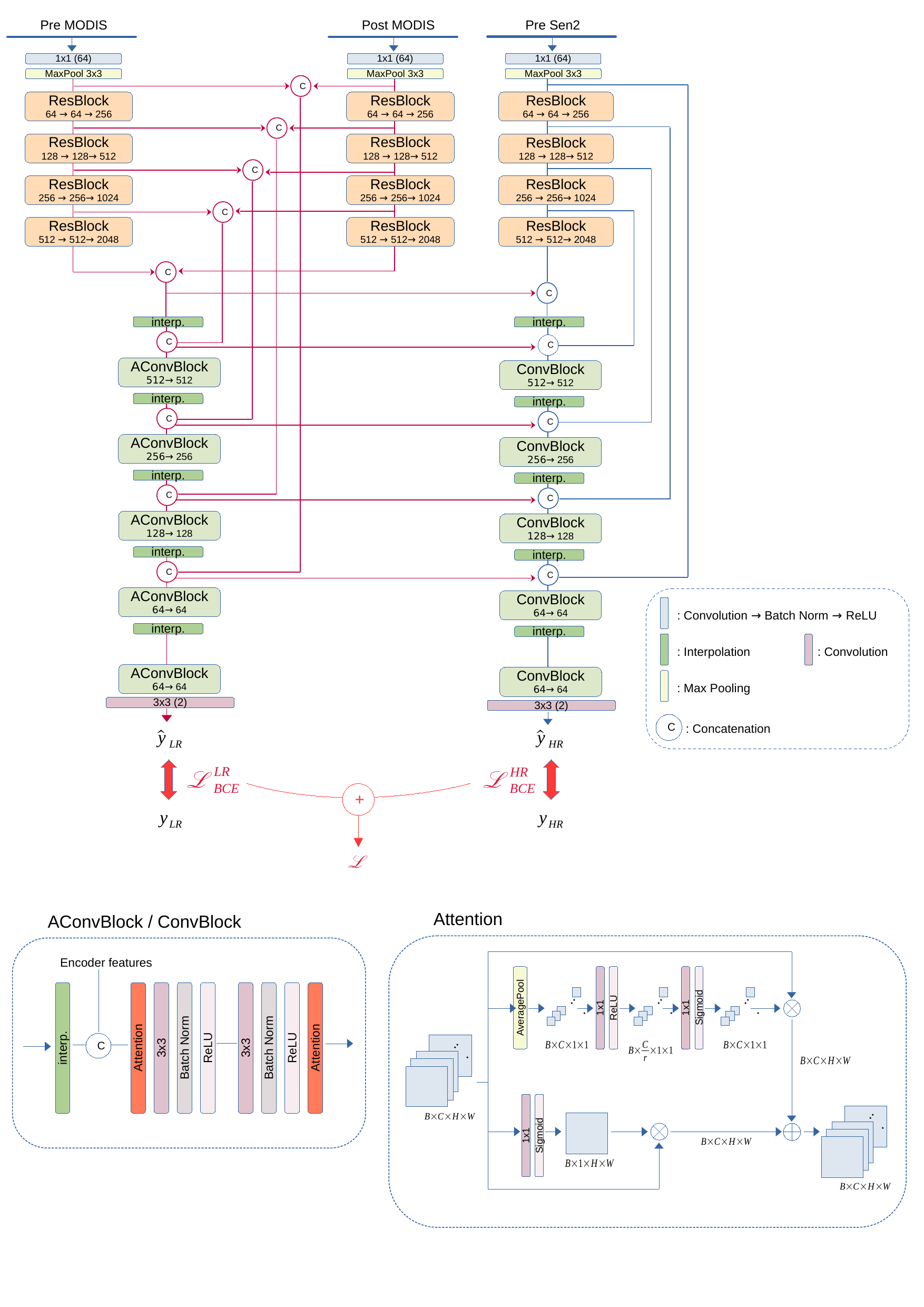}
    \caption{Overview of the BAM-MRCD model based on the BAM-CD architecture \cite{sdraka2024floga}. Numbers inside the ResBlock, ConvBlock and AConvBlock modules indicate the number of output channels for each internal convolutional layer. ConvBlock is identical to AConvBlock, without the attention module.}
    \label{fig:bam_mrcd}
\end{figure*}

\section{BAM-MRCD}
\label{sec:bam_mrcd}

The integration of $I_{t1}^{HSLT}$ and $I_{t1}^{LSHT}$ imagery into the same pipeline inevitably poses significant challenges stemming from the fundamental differences between the selected satellite sensors: Firstly, due to the resolution gap in the spatial dimension, a single $I^{LSHT}$ pixel corresponds to multiple $I^{HSLT}$ pixels, thus important information is aggregated into a single value and relevant context is minimized. Especially in the case of burn scars, the already rough perimeters are additionally blurred and become indiscernible from the surrounding structures. Producing an output at the highest spatial resolution $H \times W$ requires accurate decomposition of the input $I^{LSHT}$ pixels and retrieval of meaningful features at a sub-pixel level. Secondly, the captured bands of the two satellites are not always well aligned, since each sensor operates on somewhat different ranges of the e/m spectrum (Fig. \ref{fig:spectral_responses}). Such discrepancies may pose serious obstacles in the training process and hinder the model's ability to combine information from the two sensors in order to extract complementary information.

To alleviate the aforementioned issues, we design a novel deep learning architecture which processes the multi-source input in a parallel fashion, independently extracting meaningful features from each satellite, and at the same time learns two distinct goals which distill this complex problem into simpler subtasks. In particular, we propose BAM-MRCD (Burnt Area Mapping - MultiResolution Change Detection), a deep learning model for the task of burn scar mapping with multi-resolution, multi-source satellite input. This model is heavily based on BAM-CD \cite{sdraka2024floga}, an architecture for change detection achieving state-of-the-art results in burn scar delineation with Sentinel-2 bitemporal imagery. BAM-CD is a DeepResUnet \cite{yi_semantic_2019} model with two parallel Siamese encoders, whose weights are shared. The decoder comprises convolutional blocks (\textit{AConvBlock}) with ``squeeze-and-excitation'' attention layers which help focus on the more salient features corresponding to the burn scar.

Extending BAM-CD, we propose BAM-MRCD, a combination of two deep learning architectures trained in parallel with a deep supervision loss (Fig. \ref{fig:bam_mrcd}). In more detail, a BAM-CD takes as input the bitemporal MODIS imagery (one image in each encoder) and a UNet-like model takes as input the Sentinel-2 pre-fire image. The features produced by the BAM-CD encoders are concatenated and fed to the corresponding decoder whose internal attention module isolates the relevant MODIS features from the bitemporal imagery. These features are then concatenated with those extracted by the UNet model and passed to the UNet decoder for the final prediction. At the same time, the BAM-CD decoder produces an auxiliary prediction at the resolution of MODIS (i.e. 500m GSD), $\hat{y}_{LR}$. This coarser prediction provides an easier intermediate task for the model and guides BAM-CD towards performing change detection between the two MODIS captions, thus capturing the observed differences relevant to the wildfire before those are passed to the UNet as auxiliary information to assist the multi-resolution task and the final prediction, $\hat{y}_{HR}$.

Each prediction is compared with the corresponding ground truth target using the binary cross-entropy loss:

\begin{equation}
\label{eq:bce}
L_{BCE}(y, ~ \hat{y}) = -(y ~ log(\hat{y}) + (1 - y) ~ log(1 - \hat{y}))
\end{equation}

where $y$ is the ground truth and $\hat{y}$ is the model prediction. Following a deep supervision scheme, BAM-MRCD is trained with the following compound loss function, which involves (i) the loss between the auxiliary prediction $\hat{y}_{LR}$ and the low resolution target $y_{LR}$, and (ii) the final prediction $\hat{y}_{HR}$ and the high resolution target $y_{HR}$:

\begin{equation}
\label{eq:deep_sup}
L ~=~ L_{BCE}^{LR}(y_{LR}, ~\hat{y}_{LR}) ~+ ~ L_{BCE}^{HR}(y_{HR}, ~\hat{y}_{HR})
\end{equation}

The labels provided by FLOGA are used for the target $y_{HR}$, whereas an interpolated version of those labels are used as the low-resolution target $y_{LR}$.

\section{Experimental setup}
\label{sec:experiments}

In this section, we explore different techniques to obtain a high-resolution mapping from low-resolution (Approach 1) or multi-resolution (Approaches 2 and 3) bitemporal input. We consider several architectures proposed for CD tasks and modify their input accordingly. In particular, we select FC-EF-Diff \cite{daudt_fully_2018}, FC-EF-Conc \cite{daudt_fully_2018}, SNUNet-CD \cite{9355573}, ChangeFormer \cite{bandara_transformer-based_2022}, MLA-Net \cite{10381757} and ChangeMamba \cite{chen2024changemamba} and train them with both a low-resolution input (i.e. pre-fire and post-fire MODIS imagery) and multi-resolution input (i.e. a pre-fire Sentinel-2 and a post-fire MODIS image). For Approach 2 we also employ MM-Trans, a model specifically designed for multi-resolution change detection tasks with remote sensing data. This model employs transformers and was initially tested with various image sources and scaling factors, outperforming established change detection models. Furthermore, we experiment with DCILNet which comprises an SR module and two parallel encoders for the different resolutions. This model is also trained using a deep supervision loss on an interpolated version of the ground truth. In our experiments, we also considered SRCDNet, but it failed to converge to a local optimum, probably due to severe error propagation from the SR module that is trained separately \cite{li_dual-branch_2025}. Finally, Approach 3 is represented by our proposed model, BAM-MRCD, which is the first in our knowledge to exploit multi-resolution pre-change and low-resolution post-change input.

To establish the upper performance bound that can realistically be achieved, i.e. the ideal scenario where high resolution imagery is available for both time steps and all necessary information for a precise delineation is present, we report results for BAM-CD using solely Sentinel-2 bitemporal input. For a fair comparison, we use Sentinel-2 imagery at 60m GSD and downsample the ground truth masks accordingly.

In each experiment, we use all available bands from each sensor. However, when siamese structures are involved in the architecture, and features from both sensors are extracted by a single encoder, we opt to keep only the bands corresponding to similar frequency ranges (see Tab. \ref{tab:common_bands}) in order to obtain the same number of bands for both satellites. This is the case for FC-EF-Diff, FC-EF-Conc, SNUNet-CD, ChangeFormer, MLA-Net, ChangeMamba and DCILNet.

\subsection{Metrics}
\label{sec:metrics}

\begin{figure*}[!h]
\captionsetup[subfigure]{labelformat=empty}
\centering
\subfloat[]{\includegraphics[width=1in]{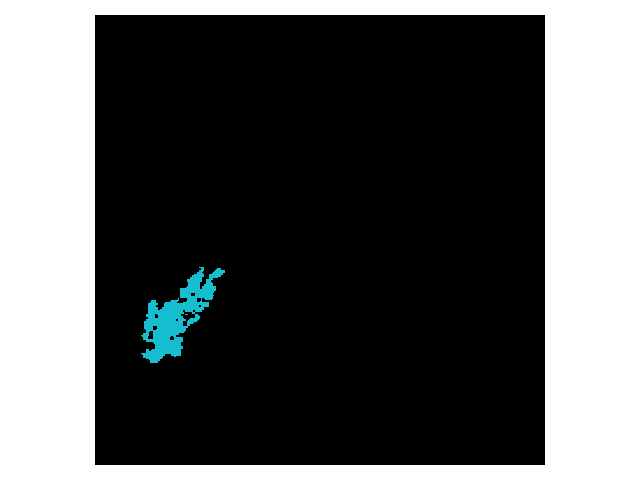}%
}
\hspace{-3.5mm}%
\subfloat[]{\includegraphics[width=1in]{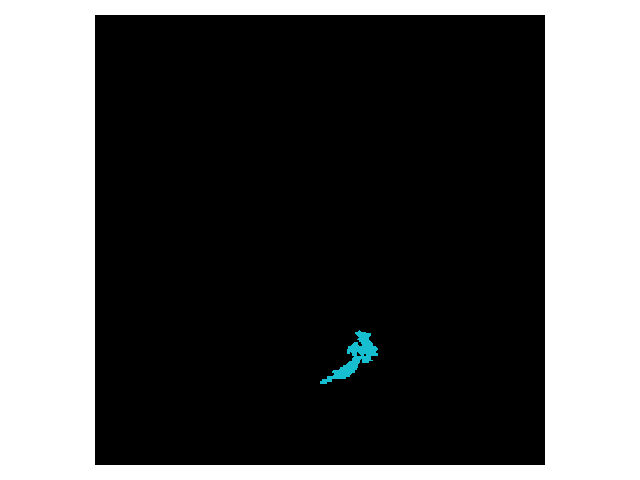}%
}
\hspace{-3.5mm}%
\subfloat[]{\includegraphics[width=1in]{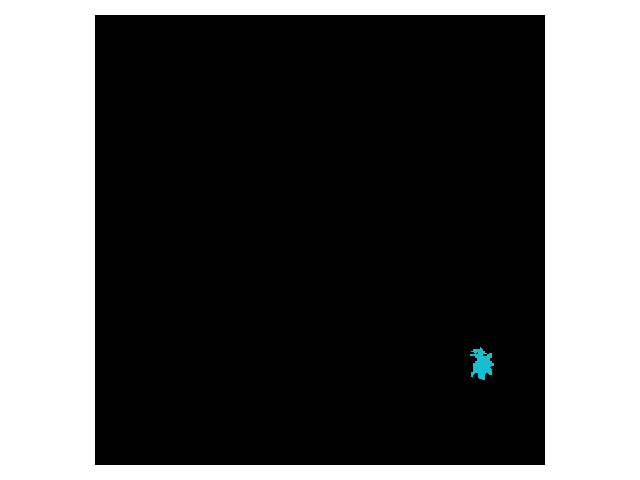}%
}
\hspace{-3.5mm}%
\subfloat[]{\includegraphics[width=1in]{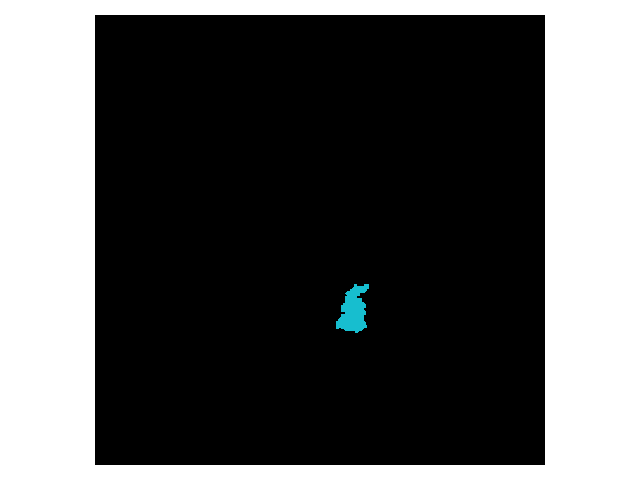}%
}
\hspace{-3.5mm}%
\subfloat[]{\includegraphics[width=1in]{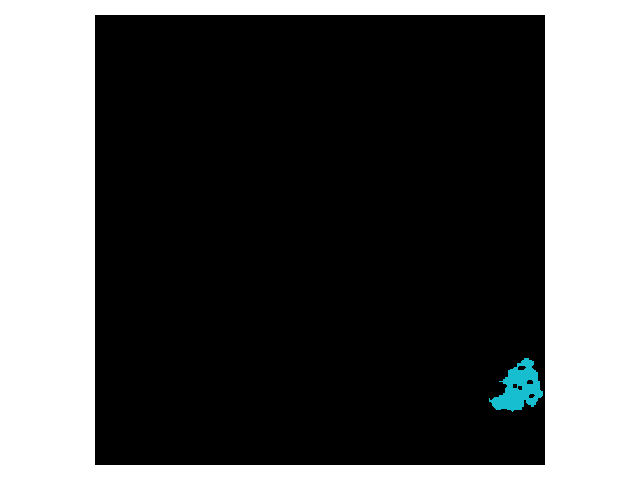}%
}

\begin{center}
\vspace*{-12ex}
\hspace*{-14cm}
Small
\end{center}
\vspace*{4ex}

\subfloat[]{\includegraphics[width=1in]{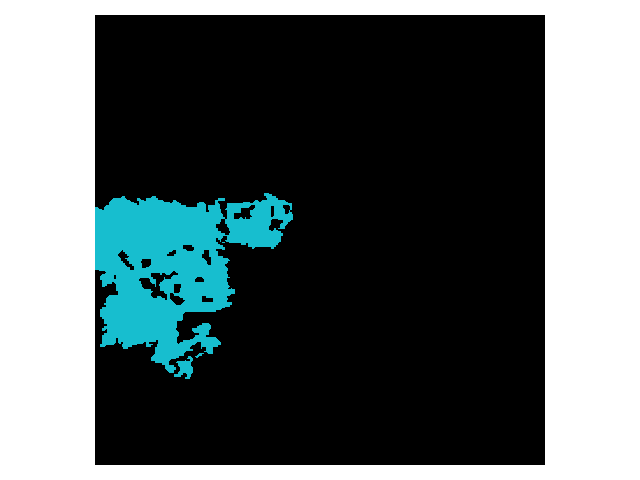}%
}
\hspace{-3.5mm}%
\subfloat[]{\includegraphics[width=1in]{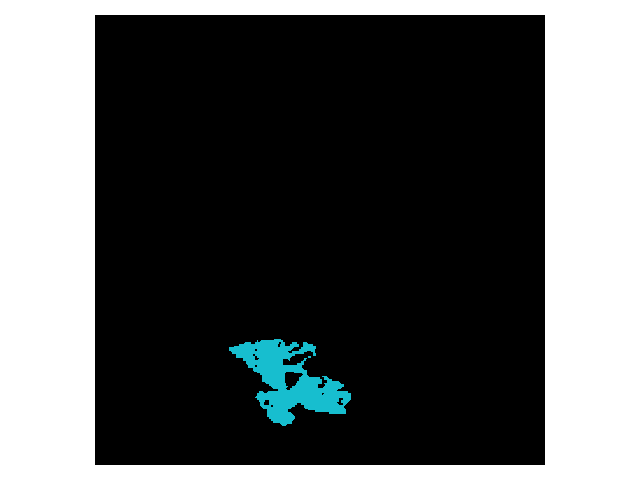}%
}
\hspace{-3.5mm}%
\subfloat[]{\includegraphics[width=1in]{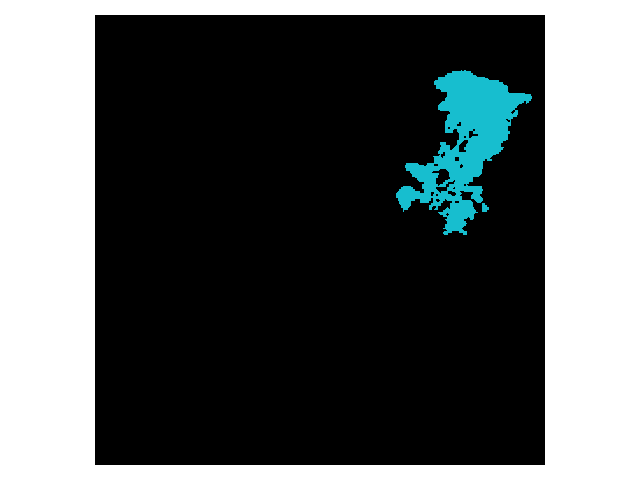}%
}
\hspace{-3.5mm}%
\subfloat[]{\includegraphics[width=1in]{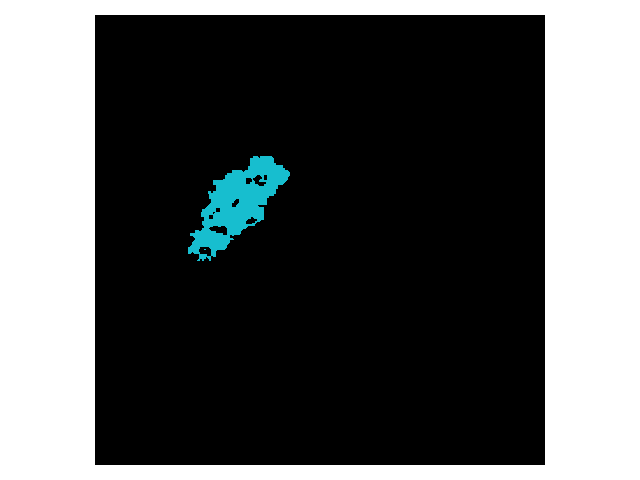}%
}
\hspace{-3.5mm}%
\subfloat[]{\includegraphics[width=1in]{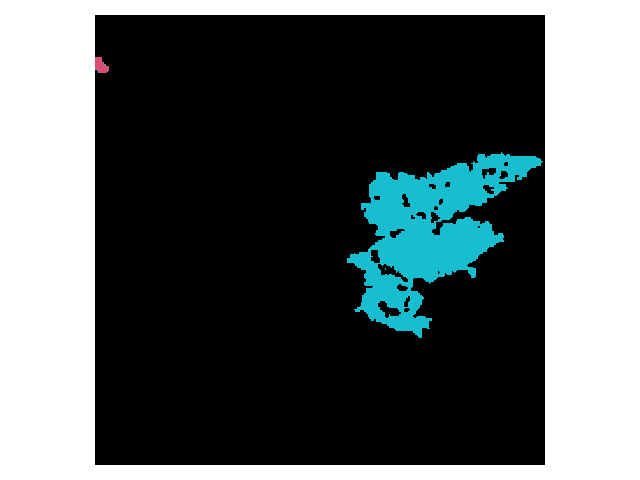}%
}

\begin{center}
\vspace*{-13ex}
\hspace*{-14cm}
Medium
\end{center}
\vspace*{5ex}

\subfloat[]{\includegraphics[width=1in]{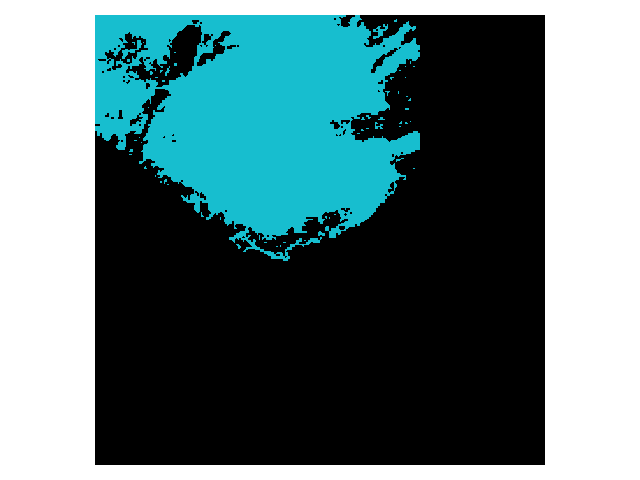}%
}
\hspace{-3.5mm}%
\subfloat[]{\includegraphics[width=1in]{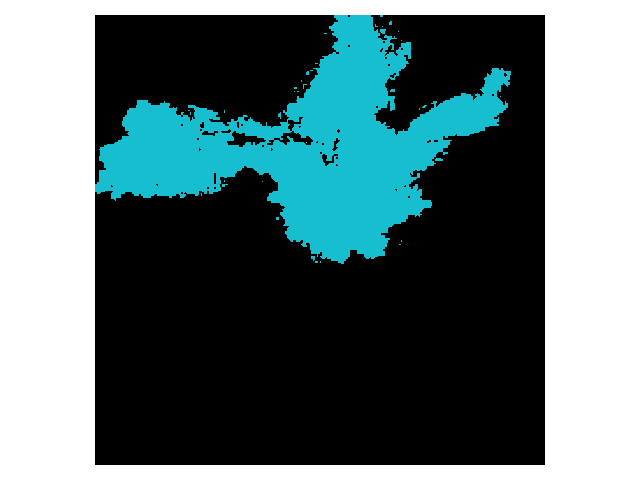}%
}
\hspace{-3.5mm}%
\subfloat[]{\includegraphics[width=1in]{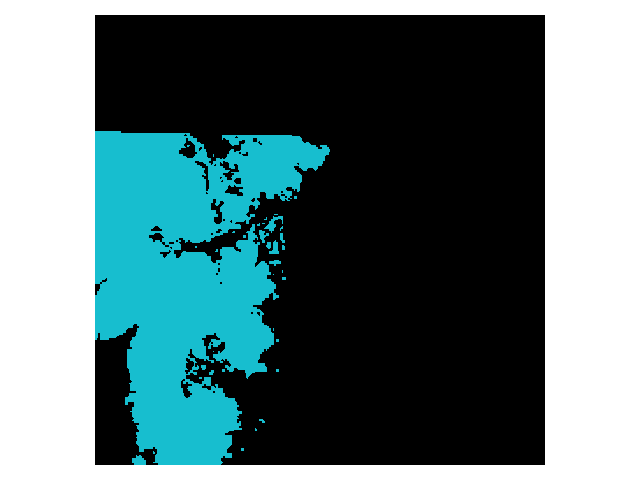}%
}
\hspace{-3.5mm}%
\subfloat[]{\includegraphics[width=1in]{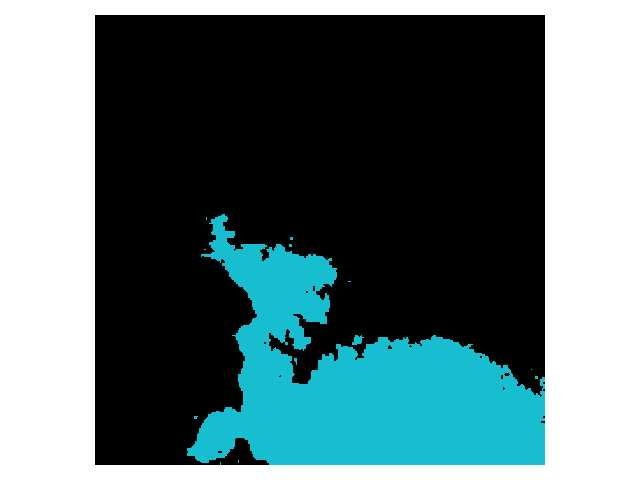}%
}
\hspace{-3.5mm}%
\subfloat[]{\includegraphics[width=1in]{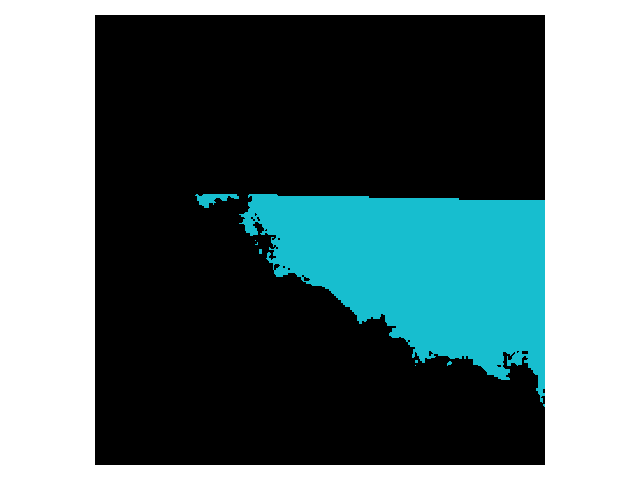}%
}

\begin{center}
\vspace*{-13ex}
\hspace*{-14cm}
Large
\end{center}
\vspace*{6ex}
\caption{Example of patch grouping into \textit{small}, \textit{medium} and \textit{large} based on the size of the burnt area.}
\label{fig:aoi_groups}
\end{figure*}

We utilise \textit{precision}, \textit{recall}, \textit{F1 score} and \textit{Intersection over Union (IoU)} as our core evaluation metrics. Nevertheless, structure-unaware metrics such as F1 score seem to be heavily affected by large detected regions which contribute more to the number of true positives and boost the final metric value. On the other hand, metrics comparing the geometric similarity of the label and the prediction, such as IoU, seem to also be saturated by large overlaps, losing sensitivity towards smaller undetected regions or inaccurate outlines. Therefore, we argue that these metrics fail to offer a good indicator of a model's robustness, especially in cases where significant variability in the size of the target area is observed. To that end, we propose a \textit{multi-scale IoU metric}, which imposes a stricter evaluation criterion and offers richer insights on varying scale delineations. We define this metric by grouping the target masks into 3 disparate groups (\textit{small}, \textit{medium}, \textit{large}) based on the number of positive (i.e. burnt) pixels and calculating a different IoU score for each of them:

\begin{equation}
IoU ~\Rightarrow ~ \left\{
\begin{array}{ll}
      IoU_{S}, ~~ if ~~ n_{pos} < th_{1} \\
      IoU_{M}, ~ if ~~ th_{1} \leq n_{pos} < th_{2} \\
      IoU_{L}, ~~ if ~~ th_{2} \leq n_{pos} \\
\end{array} 
\right.
\end{equation}

where $n_{pos}$ is the number of positive pixels in the ground truth mask, and $th_{1}$, $th_{2}$ two thresholds defining the different groups. For the FLOGA dataset, we empirically set the thresholds to $th_{1} = 0.02 * H * W$ and $th_{2} = 0.1 * H * W$, where $H$ and $W$ are the height and width of the mask, respectively. Examples of such a grouping is shown in Fig.~\ref{fig:aoi_groups}.

In addition, we report the \textit{number of events} correctly detected by the model, where at least one pixel of the prediction overlaps with the ground truth. This metric can provide useful insights on the ability of the model to retrieve information relevant to wildfire and unveil impacted regions.

\begin{table}[t!]
\caption{The number of trainable parameters (in millions) for each of the examined models.}
\label{tab:training_params}
\centering
\begin{tabular}{@{}l|l}
\toprule
 \textbf{Model} & \textbf{\makecell{\# Trainable \\ params (M)}} \\
\midrule
 FC-EF-Diff & 1.35 \\
 FC-EF-Conc & 1.5  \\
 SNUNet-CD & 12.04 \\
 ChangeFormer & 41.03 \\
 MLA-Net & 239.61 \\
 ChangeMamba & 54.0 \\
 MM-Trans & 34.68 \\
 DCILNet & 70.57 \\
 BAM-CD & 126.22 \\  
 BAM-MRCD & 340.07 \\
\bottomrule
\end{tabular}
\end{table}

\subsection{Implementation details}
\label{sec:implementation}

For all experiments we employed an NVIDIA GeForce RTX 3090 Ti GPU. All training parameters of the benchmarked models were derived from their respective publications with the exception of BAM-CD which achieved noticeably better results when trained in a pseudo-siamese fashion combined with strong data augmentations. For BAM-MRCD, we utilised the Adam optimizer with a learning rate of $10^{-4}$ and a linear learning rate scheduling. Finally, to account for any possible overfitting, in all experiments we use data augmentation which involves horizontal flipping, vertical flipping and rotation within [-15, 15] degrees, each independently applied with a 50\% probability. In Tab. \ref{tab:training_params} we report the total number of trainable parameters for all models.

\subsection{Results}
\label{sec:results}

Tab. \ref{tab:results} displays the performance of all benchmarked models. Interestingly, conventional CD methods achieve higher accuracy than specialised SR-CD models (MM-Trans and DCILNet) in this particular task, especially for small- and medium-sized changes, which establishes their ability to adapt to the cross-resolution change detection problem when directly provided with spatially aligned bitemporal input from different sources and high-resolution targets. As expected, we also observe that the best case method manages to retrieve a larger number of wildfire events, and displays superior performance on smaller-sized burn scars since it is provided with all necessary high spatial frequency details in the input. Compared with the rest of the models, it becomes evident that the F1 score fails to reflect the large performance gaps and is indeed a rather weak metric for a multi-scale segmentation setting.

In the following paragraphs, we discuss the results from our benchmark and contrast the methods of each approach to our proposed model. We also present visualizations for a number of test samples aiming to provide a qualitative assessment of the models' performance. 

\textbf{Comparison with Approach 1.} Both FC-EF-Diff and FC-EF-Conc fail to converge during the training process, which is a direct result of their shallow architecture and limited capacity. ChangeFromer fails to capture large events accurately and displays unstable performance with significantly high standard deviation on most metrics, possibly owing to the limited volume of training data which is an inhibitory factor for transformer-based architectures. Finally, SNUNet-CD, MLA-Net and ChangeMamba show the most stable performance, with the latter achieving similar or superior results when compared with the rest of the models, and managing to capture the most wildfire events in the test set, confirming the strong discriminatory capabilities of state-space models such as Mamba. However, BAM-MRCD manages to outperform all models with relatively low standard deviation. Even though our model retrieved a similar number of wildfire events as ChangeMamba, it provided a more accurate delineation and fewer misclassifications as depicted by the dedicated IoU metrics. 

\textbf{Comparison with Approach 2.} Most of the examined models seem to perform worse when provided directly with multi-resolution input, which shows their inability to contrast such vastly diverse input and extract meaningful information. Results show that the ingestion of high-resolution pre-fire imagery can potentially assist ChangeFormer in the delineation of small- and medium-sized events, as well as the stabilization of the training process (lower standard deviations). However, a notable decrease in the accuracy and stability of the mapping of large wildfires events is observed, along with a decrease in the precision and F1 scores, which implies that while the model retrieves more burnt regions, at the same time it also produces a less accurate outline including more false alarms. In addition, BAM-CD displays substantial performance gains in this approach, with increased overall accuracy in the produced delineations but with significantly more unstable training. Furthermore, the reported metrics show that SR-CD methods struggle with the delineation of small- and medium-sized burn scars, and seem to recover a rather small portion of the overall test events. In general, all models in this category fail to surpass conventional CD methods, which aligns with previous findings in the literature (e.g. \cite{chen_continuous_2023}). The performance of MM-Trans is possibly hindered by the rather limited number of positive samples during training, since transformer-based architectures require orders of magnitude more training data than convolutional-based approaches. On the other hand, even though DCILNet manages to retrieve significantly more burn scars, it is evident that it underperforms with regards to smaller sized events both in terms of absolute metric values and stability. This shortcoming can be attributed to artifacts produced by the constituent SR module which are subsequently propagated to the CD component introducing noise and misclassification errors. BAM-MRCD circumvents the need for a super-resolved input image by breaking the problem down into two sub-tasks (i.e. low-resolution CD and assisted multi-resolution CD) and exporting features directly from the LR input with a separate model in a parallel fashion.

\def\arraystretch{1.5} 
\setlength\arrayrulewidth{1pt} 
\begin{table*}[t!]
\caption{Results for the selected deep learning models across 3 distinct runs, along with the best case scenario of having both Sentinel-2 pre and post captions. The tables report the mean and standard deviation for Precision (Prec), Recall (Rec), F1-score (F1), Intersection over Union for small (IoU\textsubscript{S}), medium (IoU\textsubscript{M}) and large (IoU\textsubscript{L}) burnt areas, as well as the number of retrieved events (\# Events). All metrics refer to the positive (burnt) class.
The best value in each column is marked in \textbf{bold} and the second best is \underline{underlined}, with the exception of the best case model.}
\label{tab:results}
\centering
\resizebox{\textwidth}{!}{
\begin{tabular}{@{}lccc|ccccccc@{}}
\toprule
 & \textbf{Model} & \textbf{Pre} & \textbf{Post} & \textbf{Prec} & \textbf{Rec} & \textbf{F1} & \textbf{IoU\textsubscript{S}} & \textbf{IoU\textsubscript{M}} & \textbf{IoU\textsubscript{L}} & \textbf{\# Events} \\
\midrule
\midrule
 \multirow{7}{*}{\textbf{Approach 1}} 
 & FC-EF-Diff & MOD & MOD & 53.67 ± 34.05 & 15.08 ± 4.46 & 20.43 ± 4.73 & 0.92 ± 0.68 & 8.43 ± 8.03 & 14.9 ± 3.02 & - \\
 & FC-EF-Conc & MOD & MOD & 34.95 ± 45.79 & 5.95 ± 3.96 & 8.67 ± 8.5 & 0.87 ± 0.41 & 4.5 ± 0.82 & 5.26 ± 5.89  & -  \\ 
 & SNUNet-CD & MOD & MOD & 94.06 ± 0.77 & 88.52 ± 0.64 & \underline{91.2 ± 0.07} & 24.43 ± 0.79 & 64.58 ± 0.07 & 93.84 ± 0.22 & \underline{31 ± 1} \\ 
 & ChangeFormer & MOD & MOD & \textbf{95.24 ± 2.2} & 64.9 ± 10.76 & 76.71 ± 8.29 & 11.79 ± 5.65 & 54.23 ± 10.01 & 69.98 ± 14.0 & 19 ± 5 \\
 & MLA-Net & MOD & MOD & 93.23 ± 0.77 & 88.24 ± 0.84 & 90.66 ± 0.08 & 19.64 ± 1.79 & 64.24 ± 0.28 & 93.26 ± 0.27 & 26 ± 4 \\
 & ChangeMamba & MOD & MOD & 91.18 ± 0.39 & \underline{90.31 ± 0.27} & 90.74 ± 0.09 & \underline{26.97 ± 0.57} & 65.34 ± 0.42 & 93.17 ± 0.43 & \textbf{36 ± 2} \\
 & BAM-CD & MOD & MOD & 93.73 ± 0.27 & 88.55 ± 0.21 & 91.07 ± 0.06 & 22.48 ± 0.94 & 65.08 ± 0.35 & 93.87 ± 0.24 & \underline{31 ± 2} \\
 \midrule
 \midrule
\multirow{9}{*}{\textbf{Approach 2}}
 & FC-EF-Diff & Sen-2 & MOD & 4.28 ± 1.17 & 34.69 ± 19.0 & 7.03 ± 1.53 & 0.59 ± 0.07 & 5.28 ± 0.45 & 6.59 ± 1.47 & - \\
 & FC-EF-Conc & Sen-2 & MOD & 6.44 ± 5.21 & 25.71 ± 8.64 & 8.09 ± 4.27 & 0.62 ± 0.05 & 4.53 ± 0.63 & 6.99 ± 3.12 & - \\
 & SNUNet-CD & Sen-2 & MOD & 93.67 ± 1.16 & 88.78 ± 0.52 & 91.15 ± 0.4 & 22.74 ± 2.42 & 60.2 ± 0.52 & \underline{94.75 ± 0.39} & \underline{31 ± 3} \\
 & ChangeFormer & Sen-2 & MOD & 92.41 ± 1.96 & 65.14 ± 14.28 & 75.68 ± 10.59 & 21.53 ± 1.29 & 57.99 ± 3.27 & 60.6 ± 22.74 & 25 ± 2 \\ 
 & MLA-Net & Sen-2 & MOD & 92.77 ± 5.31 & 83.01 ± 1.82 & 87.48 ± 1.47 & 8.94 ± 0.97 & 55.61 ± 3.51 & 90.66 ± 1.52 & 20 ± 5 \\
  & ChangeMamba & Sen-2 & MOD & 92.53 ± 1.0 & 89.52 ± 1.02 & 90.99 ± 0.07 & 24.44 ± 2.6 & 65.18 ± 1.7 & 93.8 ± 0.28 & \underline{31 ± 3} \\
  & BAM-CD & Sen-2 & MOD & 91.77 ± 0.99 & \textbf{90.49 ± 0.23} & 91.12 ± 0.38 & 24.84 ± 2.1 & \textbf{65.61 ± 1.37} & 93.78 ± 0.64 & \underline{31 ± 2} \\
 \cmidrule(lr){2-11}
 & MM-Trans & Sen-2 & MOD & 86.32 ± 2.8 & 85.18 ± 1.14 & 85.70 ± 0.89 & 6.51 ± 0.72 & 21.4 ± 0.98 & 51.51 ± 6.27 & 12 ± 2 \\
 & DCILNet & Sen-2 & MOD & 91.0 ± 1.65 & 65.73 ± 9.78 & 75.81 ± 5.95 & 2.7 ± 2.56 & 35.93 ± 16.46 & 73.04 ± 7.78 & 19 ± 5 \\
 \midrule
 \midrule
  \textbf{Approach 3} & \textbf{BAM-MRCD} & Sen-2/MOD & MOD & \underline{94.56 ± 0.44} & 89.69 ± 0.39 & \textbf{92.06 ± 0.02} & \textbf{27.25 ± 0.54} & \underline{65.46 ± 0.89} & \textbf{95.4 ± 0.14} & \textbf{36 ± 2} \\
 \hline
 \hline
 \rowcolor[HTML]{F4F5F6}\global\setlength\arrayrulewidth{0.4pt} 
  \textbf{Best case} & BAM-CD & Sen-2 & Sen-2 & 94.19 ± 2.21 & 92.46 ± 0.67 & 93.3 ± 0.76 & 46.35 ± 3.33 & 70.66 ± 0.87 & 95.1 ± 0.76 & 46 ± 1 \\
\hline
\end{tabular}
}
\end{table*}
\setlength\arrayrulewidth{0.4pt} 

\begin{figure*}[!t]
\captionsetup[subfloat]{captionskip=20pt}
\centering
\subfloat[]{\includegraphics[width=3.1in]{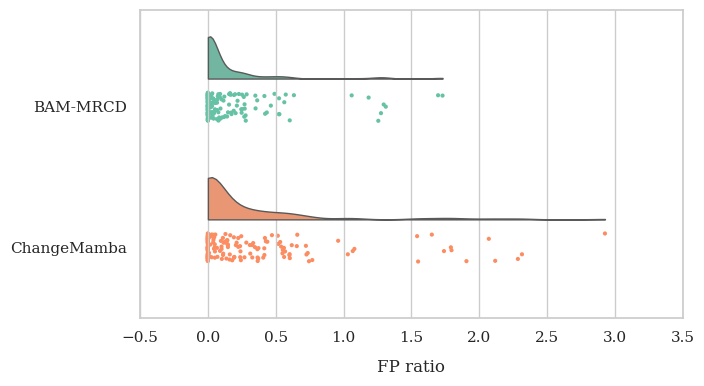}%
\label{fig:fp_ratios}}
\hspace{3.5mm}%
\subfloat[]{\includegraphics[width=3.1in]{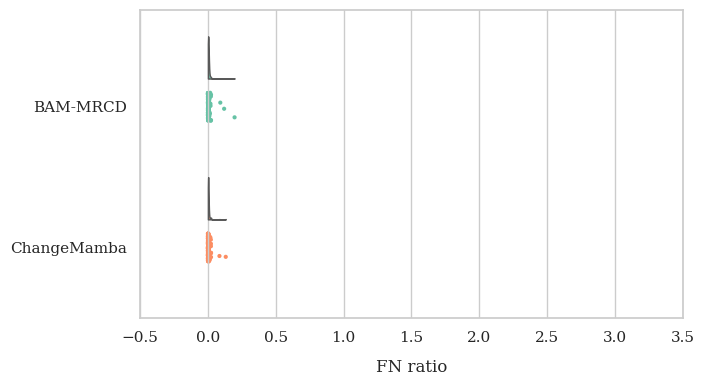}%
\label{fig:fn_ratios}}
\caption{Distributions of (a) the false positive ratio and (b) the false negative ratio for BAM-MRCD and ChangeMamba (Approach 1) in the test set. Each ratio is defined by the number of false positives or false negatives over the total number of positives or negatives, respectively.}
\label{fig:fn_fp_ratios}
\end{figure*}

\begin{figure*}[htbp]
    \centering
    \begin{tabular}{>{\raggedleft}m{0.2cm} m{14.5cm}}
        (a) &
        \includegraphics[width=\linewidth]{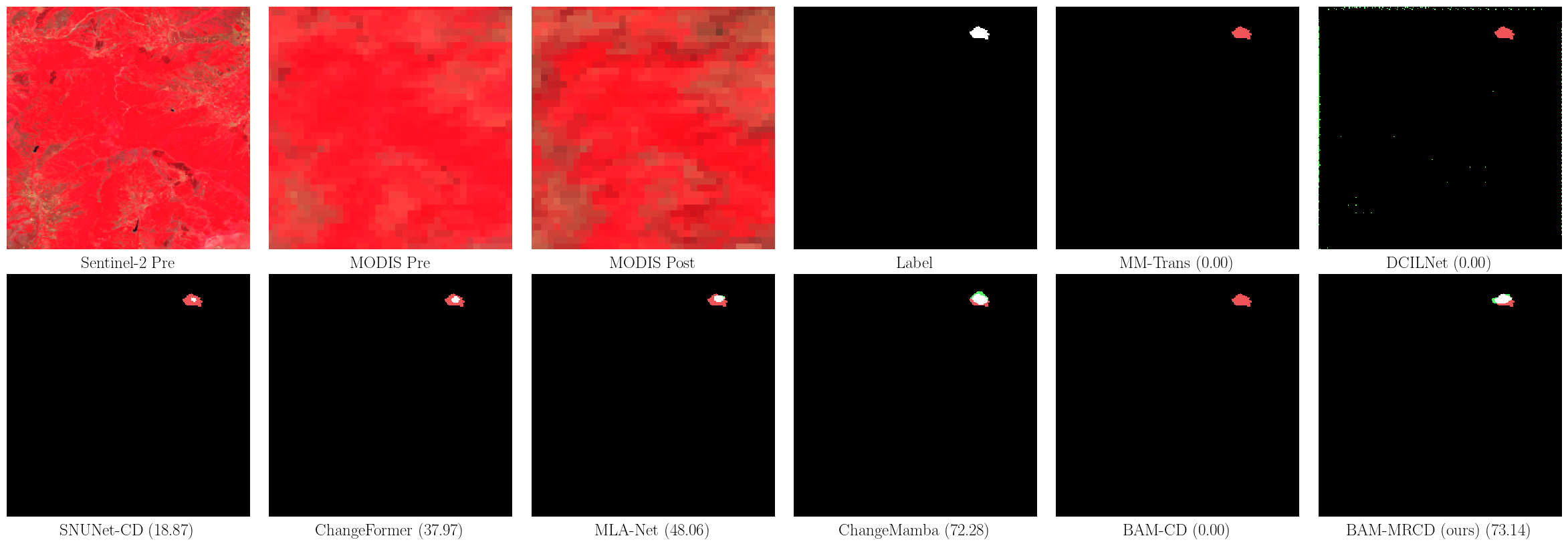} \\
        (b) &
        \includegraphics[width=\linewidth]{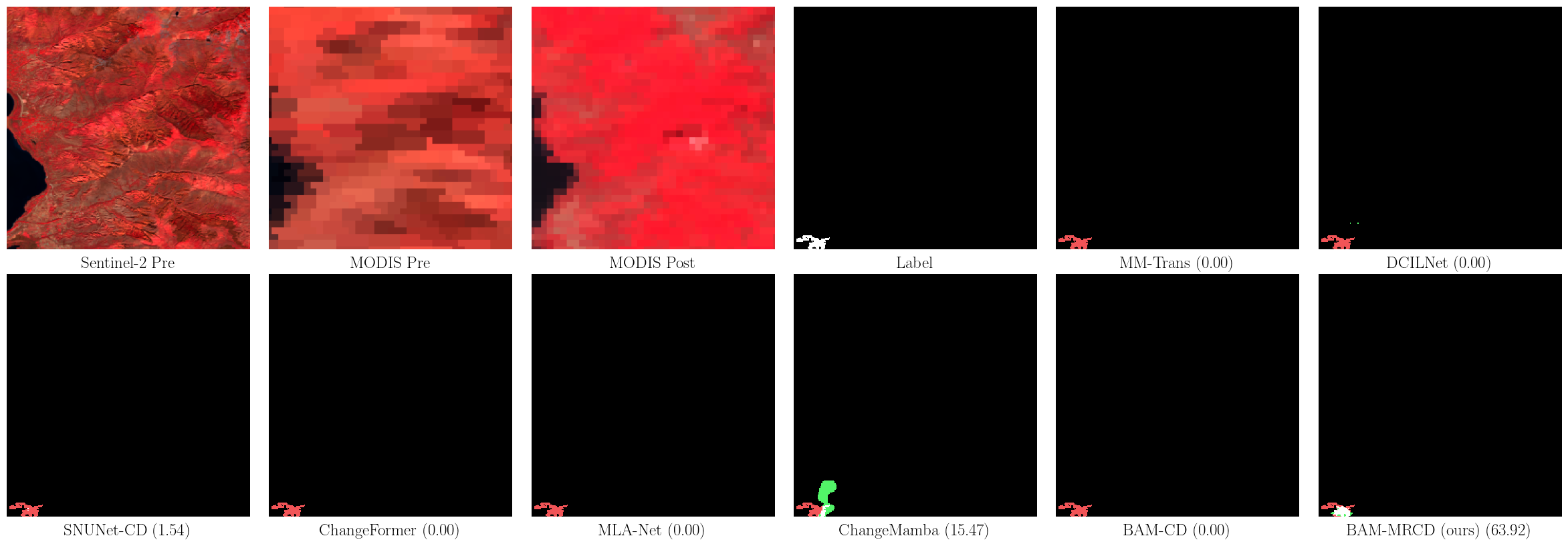} \\
        (c) &
        \includegraphics[width=\linewidth]{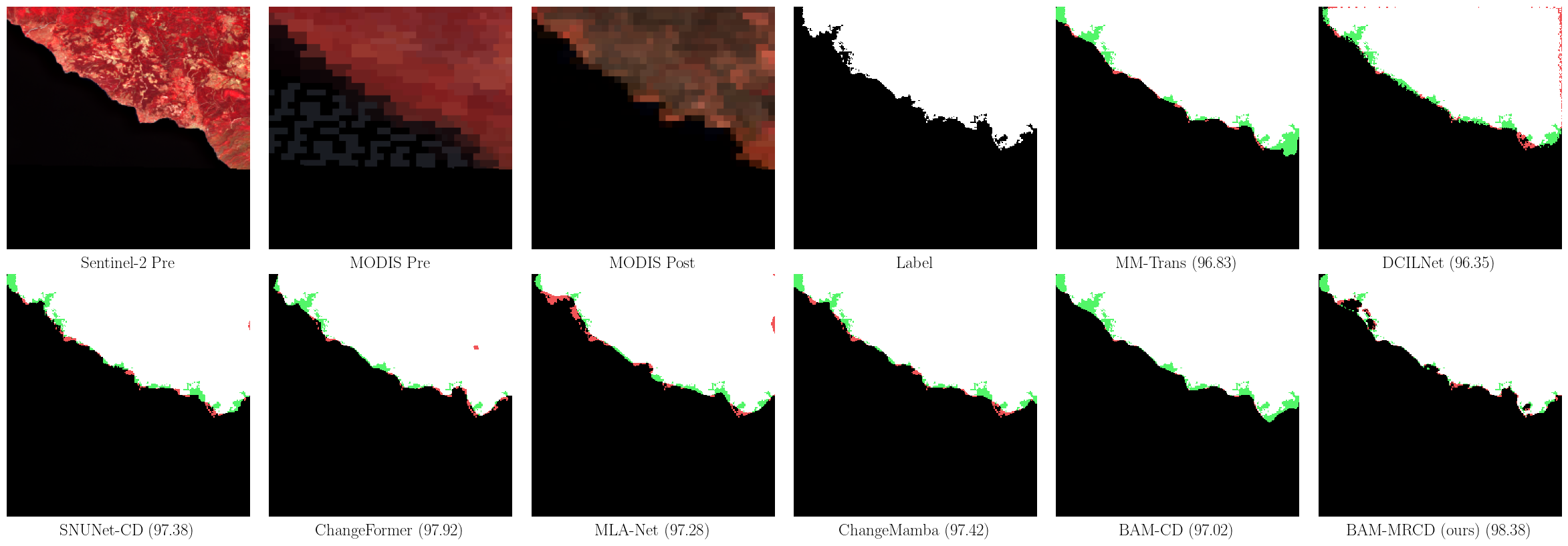} \\
        (d) &
        \includegraphics[width=\linewidth]{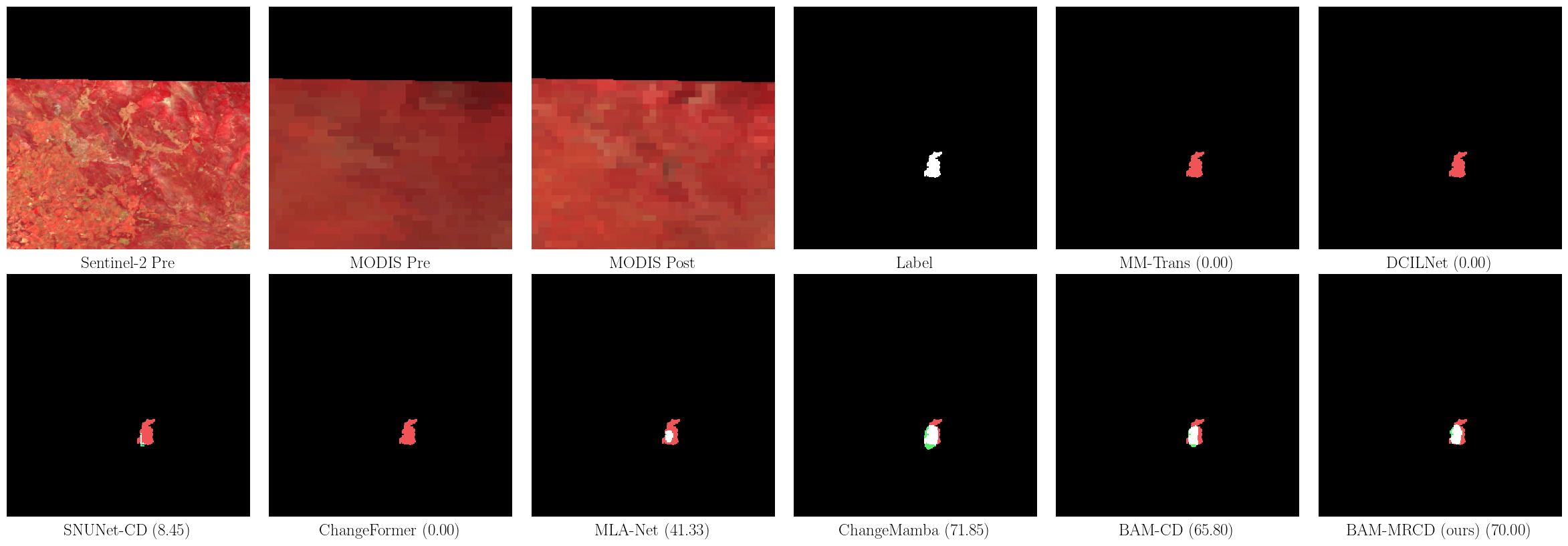}
    \end{tabular}
    \caption{Sample predictions for burn scars of various sizes in the test set. Numbers in parentheses indicate the F1-score. Satellite imagery is plotted as NIR-Red-Green composites. False negatives are indicated by \textcolor{red}{red} colour, false positives by \textcolor{green}{green} and true positives by \textpdfrender{
    TextRenderingMode=FillStroke,
    LineWidth=.2pt,
    FillColor=white,
    }{white}. Total burnt area visible in patch: (a) 52.63 ha, (b) 60.52 ha, (c) 7515.15 ha, (d) 94.94 ha.}
    \label{fig:results_tp}
\end{figure*}

\begin{figure*}[htbp]
    \centering
    \begin{tabular}{>{\raggedleft}m{0.2cm} m{14.5cm}}
        (a) &
        \includegraphics[width=\linewidth]{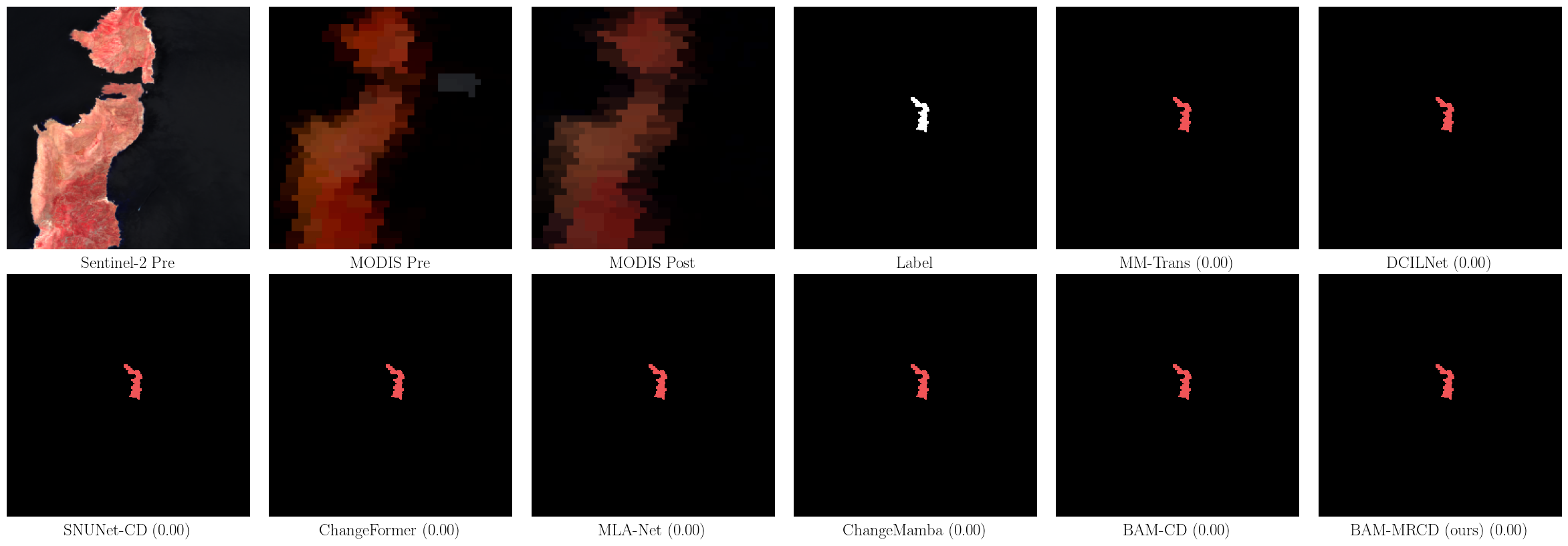} \\
        (b) &
        \includegraphics[width=\linewidth]{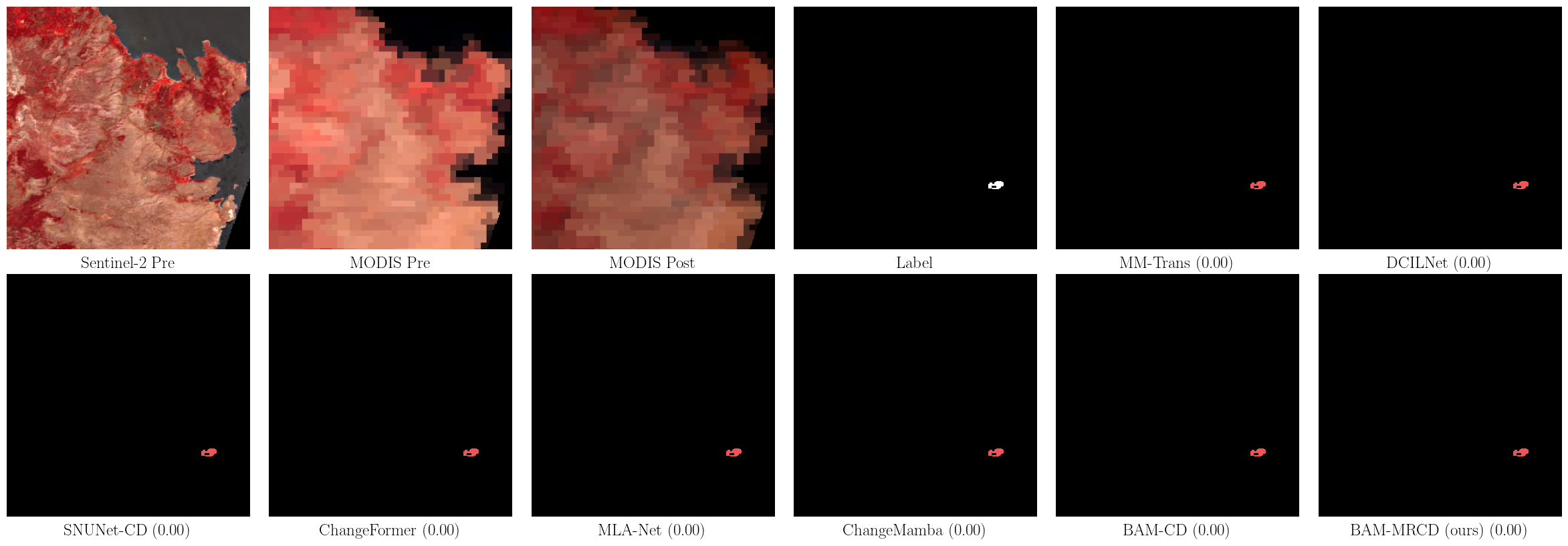} \\
        (c) &
        \includegraphics[width=\linewidth]{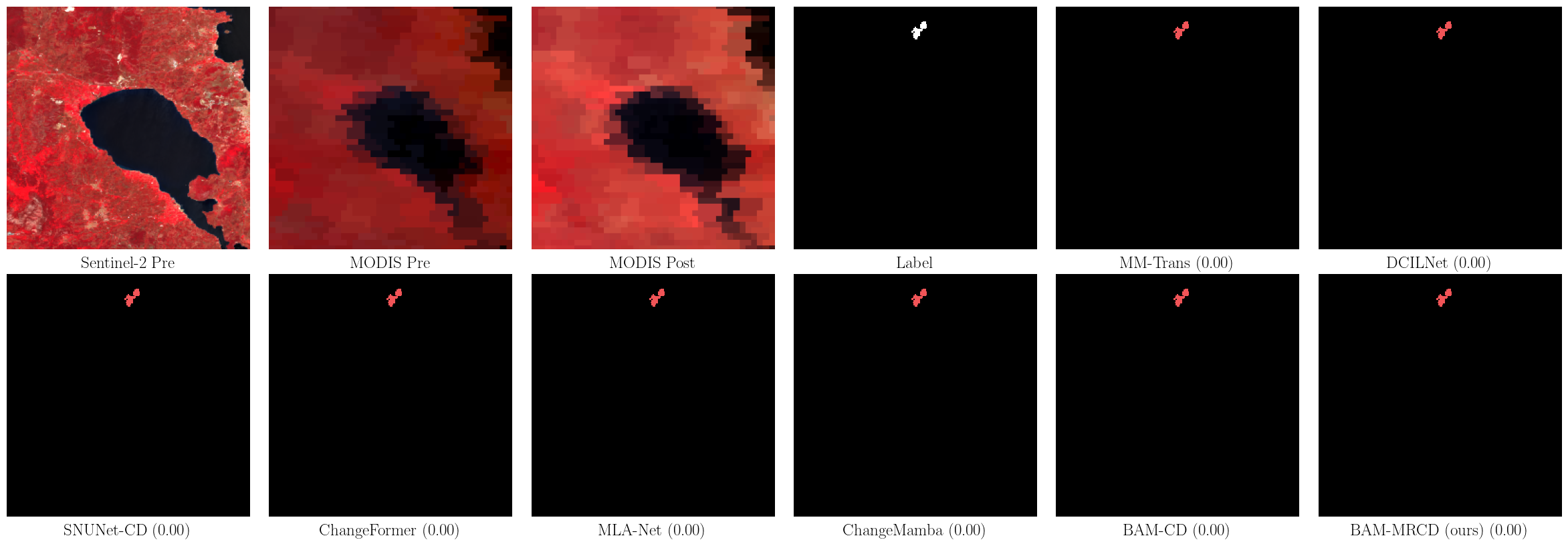} \\
        (d) &
        \includegraphics[width=\linewidth]{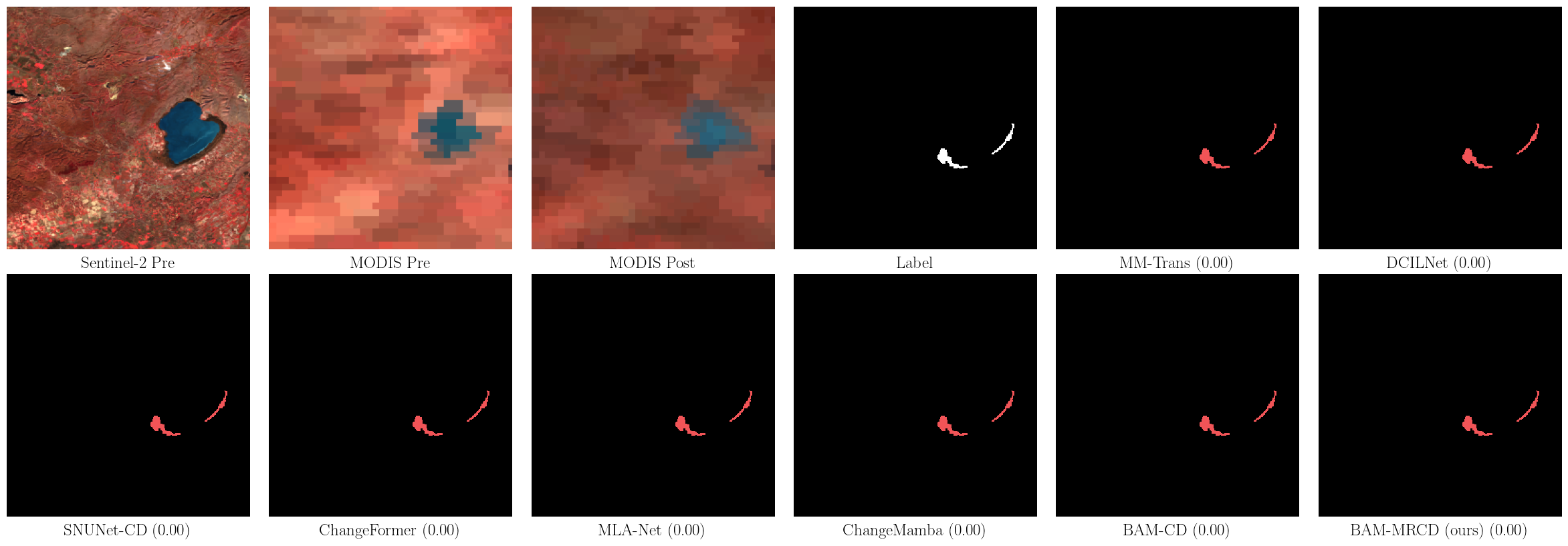}
    \end{tabular}
    \caption{Sample predictions for failure cases in the test set. Numbers in parentheses indicate the F1-score. Satellite imagery is plotted as NIR-Red-Green composites. False negatives are indicated by \textcolor{red}{red} colour, false positives by \textcolor{green}{green} and true positives by \textpdfrender{
    TextRenderingMode=FillStroke,
    LineWidth=.2pt,
    FillColor=white,
    }{white}. Total burnt area visible in patch: (a) 86.59 ha, (b) 27.39 ha, (c) 27.74 ha, (d) 72.27 ha.}
    \label{fig:results_fn}
\end{figure*}

\textbf{Qualitative results.} In Fig. \ref{fig:results_tp} and \ref{fig:results_fn} we present a number of test samples and the predictions of the models with the best reported metrics. In particular, we select SNUNet-CD, MLA-Net and ChangeMamba of Approach 1, and ChangeFormer, MM-Trans, DCILNet and BAM-CD of Approach 2. Fig. \ref{fig:results_tp} displays a number of representative cases from the test set where our model achieved superior performance. In particular, BAM-MRCD managed to retrieve and delineate small-sized burn scars that were completely or partially missed by all of the other models (Fig. \ref{fig:results_tp}a,b). At the same time, our proposed model produced a highly accurate mapping of large-sized burn scars (Fig. \ref{fig:results_tp}c), close to the provided high resolution labels, maintaining the complex outline of the affected area and limiting the number of false predictions. In addition, Fig. \ref{fig:results_tp}d shows a rather interesting case, where ChangeMamba reportedly achieved a higher F1 score, but the mapping produced by our model respects the borders of the ground truth mask and no unburnt land is misclassified. Overall, this qualitative analysis indicates that BAM-MRCD is more strongly inclined towards preserving the boundaries of the burn scar, even at the expense of slightly underestimating the total affected area. Fig. \ref{fig:fn_fp_ratios} confirms this hypothesis by showcasing the false positive and false negative ratios for the two most robust models in the benchmark, ChangeMamba (Approach 1) and BAM-MRCD. These ratios are defined as the number of false positives/negatives over the total number of positives/negatives per event, respectively. We observe that the false positive ratio distribution for BAM-MRCD is more concentrated around values closer to zero with a much shorter tail towards larger values, as compared with that of ChangeMamba, highlighting the latter's tendency to overestimate the affected regions (Fig. \ref{fig:fp_ratios}). Fig. \ref{fig:fn_ratios} showcases the reverse phenomenon for the false negative ratio, with BAM-MRCD missing more burnt pixels than ChangeMamba, albeit by a smaller margin than in the false positive case.

\textbf{Missed events.} After further investigating each missed event in the test set, we observed that all examined models failed to retrieve burn scars that spanned across less than two pixels of MODIS (i.e. approximately $<$ 90 ha) across at least one dimension. Examples of such cases are given in Fig. \ref{fig:results_fn}. As also highlighted in \cite{hall2016modis}, such small, and often partially burnt, areas provide a rather weak spectral signal compared to adjacent unburnt land in the low resolution settings of the MODIS sensor. This reduced contrast in spectral characteristics among different field conditions (such as burnt, residue-covered, plowed or bare soil) makes it even more challenging to retrieve accurate information on the burnt land, and comparison with pre-fire imagery yields no significant change.

\section{Ablation study}
\label{sec:ablation}

In order to assess the importance of each architectural element in BAM-MRCD, we perform an ablation analysis considering: (i) Siamese (S) vs. Pseudo-siamese (PS) scheme in the BAM-CD sub-module, (ii) the inclusion of attention in the BAM-CD decoder ($A_{LR}$), (iii) the inclusion of attention in the UNet decoder ($A_{HR}$). Tab. \ref{tab:ablation} reports the performance of our model with different combinations of the above parameters. We observe that $A_{LR}$ boosts the model's ability to retrieve smaller sized burn scars, which establishes the potential of attention mechanisms to isolate the most salient information in various scales. However, the addition of an extra attention layer in the UNet decoder seems to destabilize performance by increasing the standard deviation between different runs and decreasing the average of most metric values. This behaviour may indicate that the most salient features from the two decoders cannot be directly combined due to high discrepancies between the different sensors (i.e. spectral and spatial resolution gaps). On the other hand, combining all available information from the high resolution imagery along with the most important features from the low resolution change detection subtask proves to be the most effective way to isolate the underlying change while at the same time preserving the high frequency details of the burn scar boundary.
Our experiments also showed that a pseudo-siamese setting increases the flexibility and capacity of the model, providing a strong advantage in this highly complex multi-resolution task, and assists in stabilizing the model's performance across runs (as depicted by the smaller standard deviation).

\begin{table*}[ht!]
\caption{Ablation study for BAM-MRCD across 3 distinct runs. The tables report the mean and standard deviation for Precision (Prec), Recall (Rec), F1-score (F1), Intersection over Union for small (IoU\textsubscript{S}), medium (IoU\textsubscript{M}) and large (IoU\textsubscript{L}) burnt areas, as well as the number of retrieved events (\# Events). S and PS stand for siamese and pseudo-siamese respectively, $A_{LR}$ for attention in the BAM-CD decoder and $A_{HR}$ for attention in the UNet decoder. All metrics refer to the positive (burnt) class.}
\label{tab:ablation}
\centering
\resizebox{\textwidth}{!}{
\begin{tabular}{@{}ccc|ccccccc@{}}
\toprule
 \textbf{Encoder weights} & $\bm{A_{LR}}$ & $\bm{A_{HR}}$ & \textbf{Prec} & \textbf{Rec} & \textbf{F1} & \textbf{IoU\textsubscript{S}} & \textbf{IoU\textsubscript{M}} & \textbf{IoU\textsubscript{L}} & \textbf{\# Events} \\
\midrule
 S &  &  & 94.52 ± 0.57 & 89.12 ± 0.56 & 91.74 ± 0.19 & 24.2 ± 2.29 & 63.78 ± 0.79 & 95.15 ± 0.15 & 33.33 ± 3.09 \\
 PS &  &  & \underline{94.7 ± 0.34} & 88.71 ± 0.35 & 91.6 ± 0.05 & 21.2 ± 1.21 & 63.05 ± 0.33 & 95.15 ± 0.03 & 30.33 ± 2.36 \\
 S & \checkmark &  & 94.44 ± 0.54 & \underline{89.65 ± 0.38} & \underline{91.99 ± 0.12} & 26.85 ± 1.71 & \underline{65.31 ± 0.84} & \underline{95.28 ± 0.22} & \textbf{36.67 ± 0.94} \\
 PS & \checkmark &  & 94.56 ± 0.44 & \textbf{89.69 ± 0.39} & \textbf{92.06 ± 0.02} & \underline{27.25 ± 0.54} & \textbf{65.46 ± 0.89} & \textbf{95.4 ± 0.14} & \underline{36.33 ± 1.25} \\
 S &  & \checkmark & 94.47 ± 0.53 & 88.9 ± 0.33 & 91.6 ± 0.11 & 26.59 ± 1.31 & 47.44 ± 2.78 & 94.93 ± 0.21 & 30.67 ± 1.25 \\
 PS &  & \checkmark & 94.03 ± 0.91 & 88.97 ± 0.23 & 91.43 ± 0.32 & 23.02 ± 1.58 & 45.27 ± 0.54 & 94.68 ± 0.63 & 31.67 ± 2.05 \\
 S & \checkmark & \checkmark & \textbf{95.04 ± 0.42} & 88.32 ± 0.12 & 91.56 ± 0.26 & 23.7 ± 1.86 & 48.19 ± 2.86 & 95.16 ± 0.64 & 31.0 ± 1.41 \\ 
 PS & \checkmark & \checkmark & 94.49 ± 0.36 & 89.28 ± 0.33 & 91.81 ± 0.15 & \textbf{28.86 ± 1.74} & 49.48 ± 2.14 & 95.13 ± 0.17 & 32.33 ± 0.47 \\
\bottomrule
\end{tabular}
}
\end{table*}

\section{Evaluation on new events}
\label{sec:new_events}

\begin{figure*}[!t]
\captionsetup[subfloat]{captionskip=10pt}
\centering
\subfloat[]{\includegraphics[width=2in]{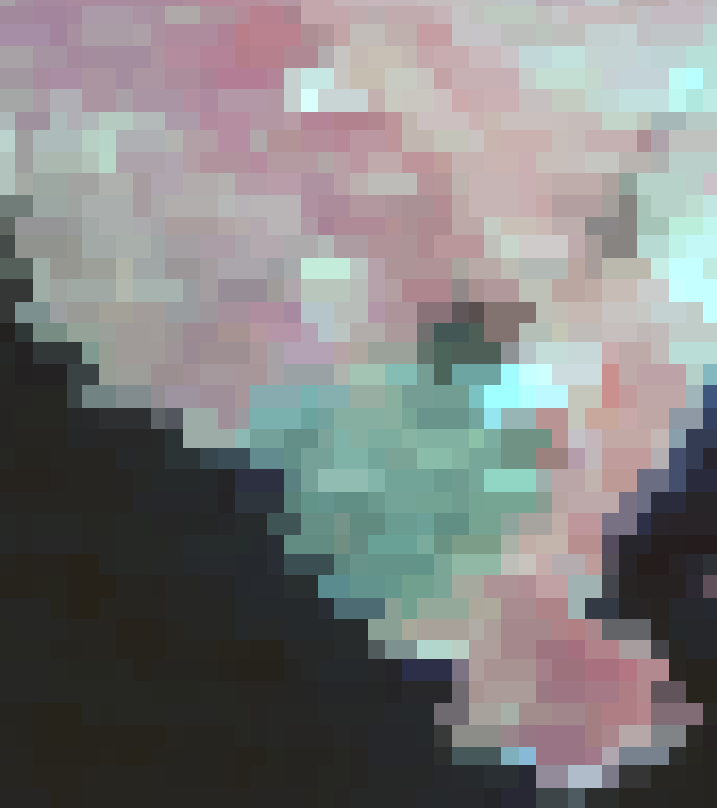}%
\label{res:832_mod}}
\hspace{3.5mm}%
\subfloat[]{\includegraphics[width=2in]{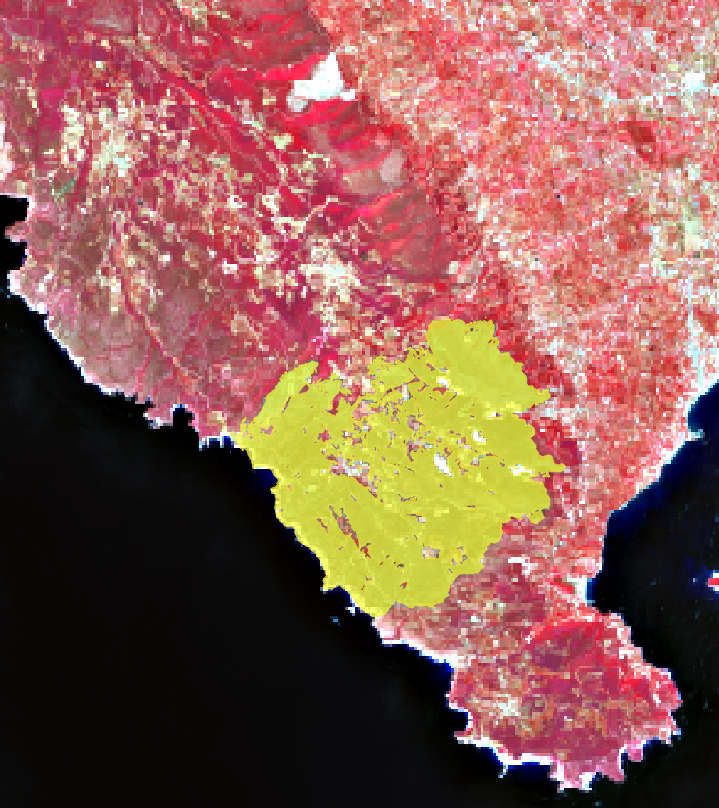}%
\label{res:832_cems}}
\hspace{3.5mm}%
\subfloat[]{\includegraphics[width=2in]{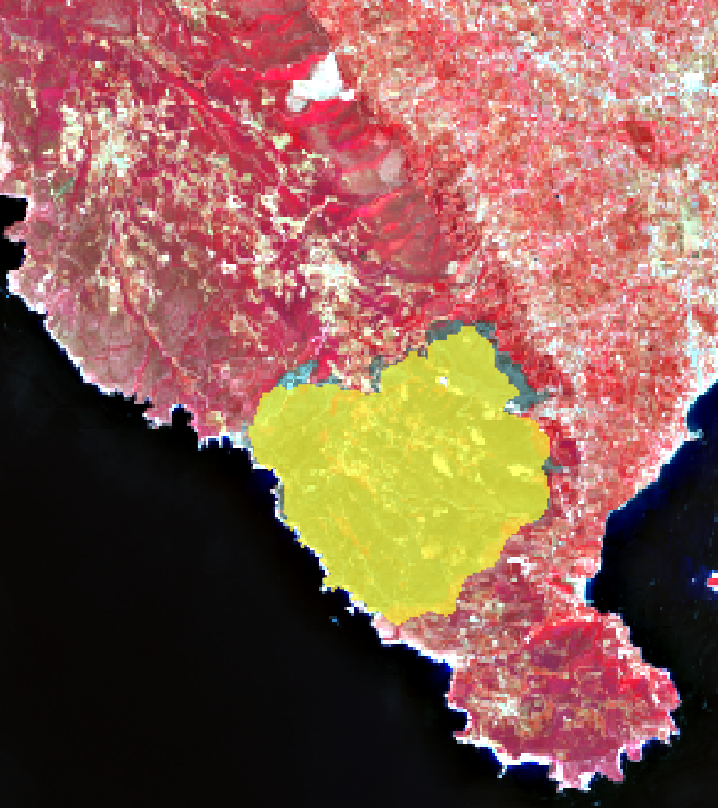}%
\label{res:832_pred}}
\caption{Post-fire Sentinel-2 caption for a wildfire in Zakynthos island, Greece, along with (a) MODIS post-event image (14/8/2025), (b) CEMS delineation on 14/8/2025, (c) BAM-MRCD delineation. (b) and (c) include the corresponding Sentinel-2 image as background. BAM-MRCD utilized a MODIS post-fire caption on 14/8/2025, whereas the first available Sentinel-2 caption was on 15/8/2025. The wildfire affected approximately 2470.6 ha. The model has managed to correctly retrieve the outline of the affected area.}
\label{res:832}
\end{figure*}

\begin{figure*}[!t]
\captionsetup[subfloat]{captionskip=10pt}
\centering
\subfloat[]{\includegraphics[width=2.2in]{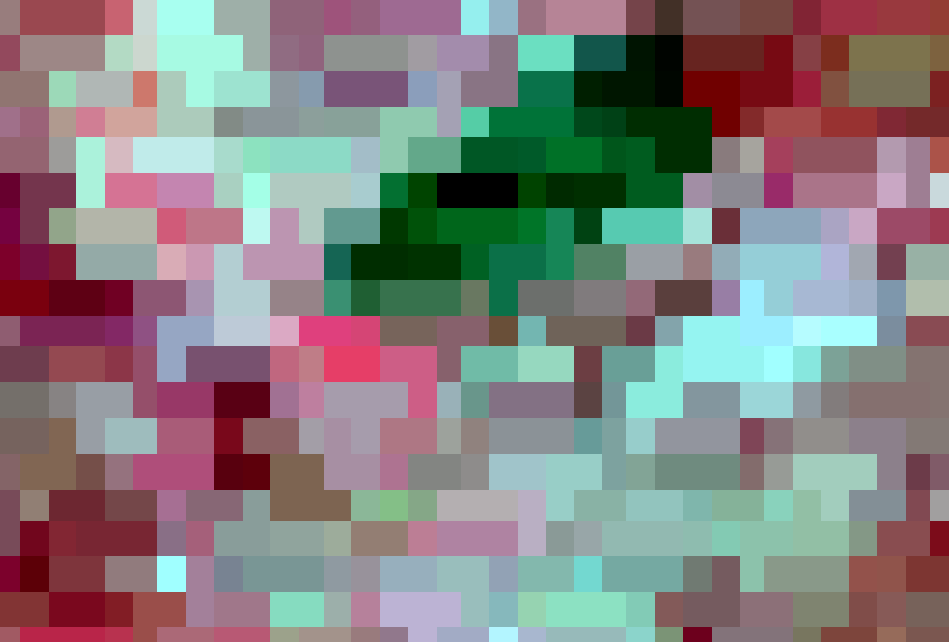}%
\label{res:829_mod}}
\hspace{3.5mm}%
\subfloat[]{\includegraphics[width=2.2in]{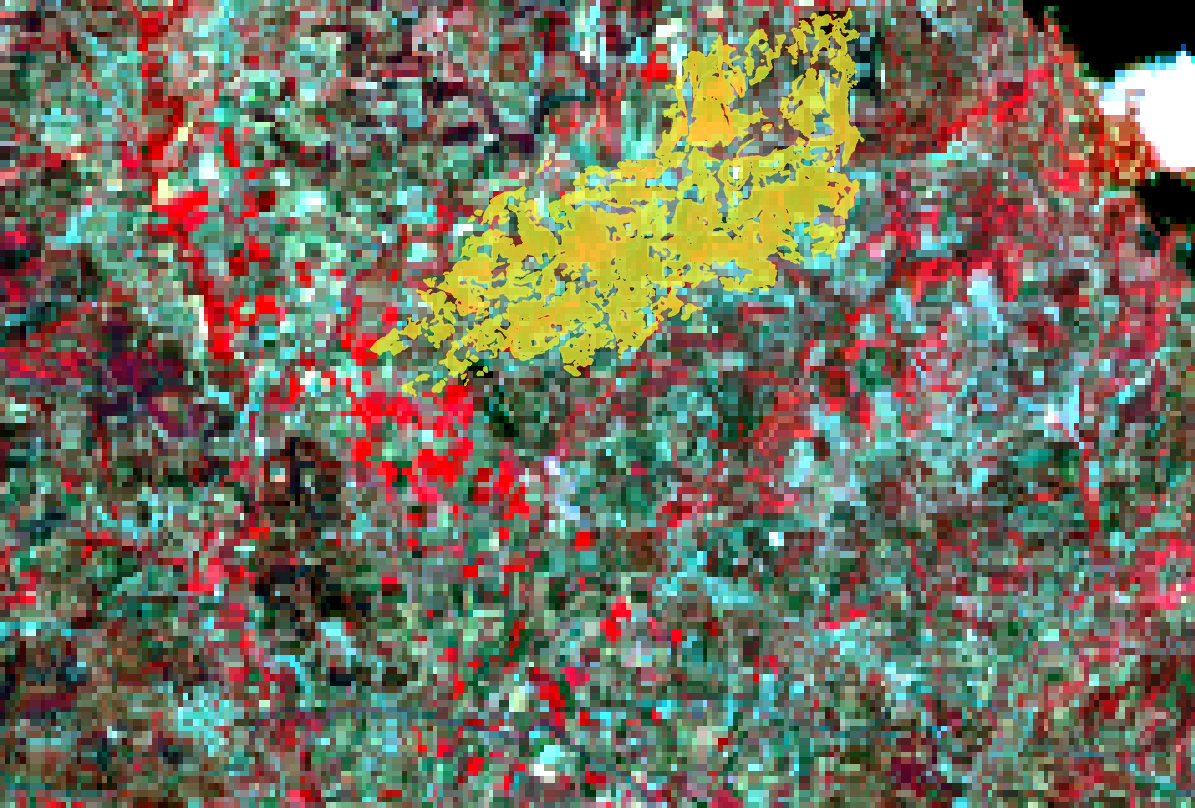}%
\label{res:829_cems}}
\hspace{3.5mm}%
\subfloat[]{\includegraphics[width=2.2in]{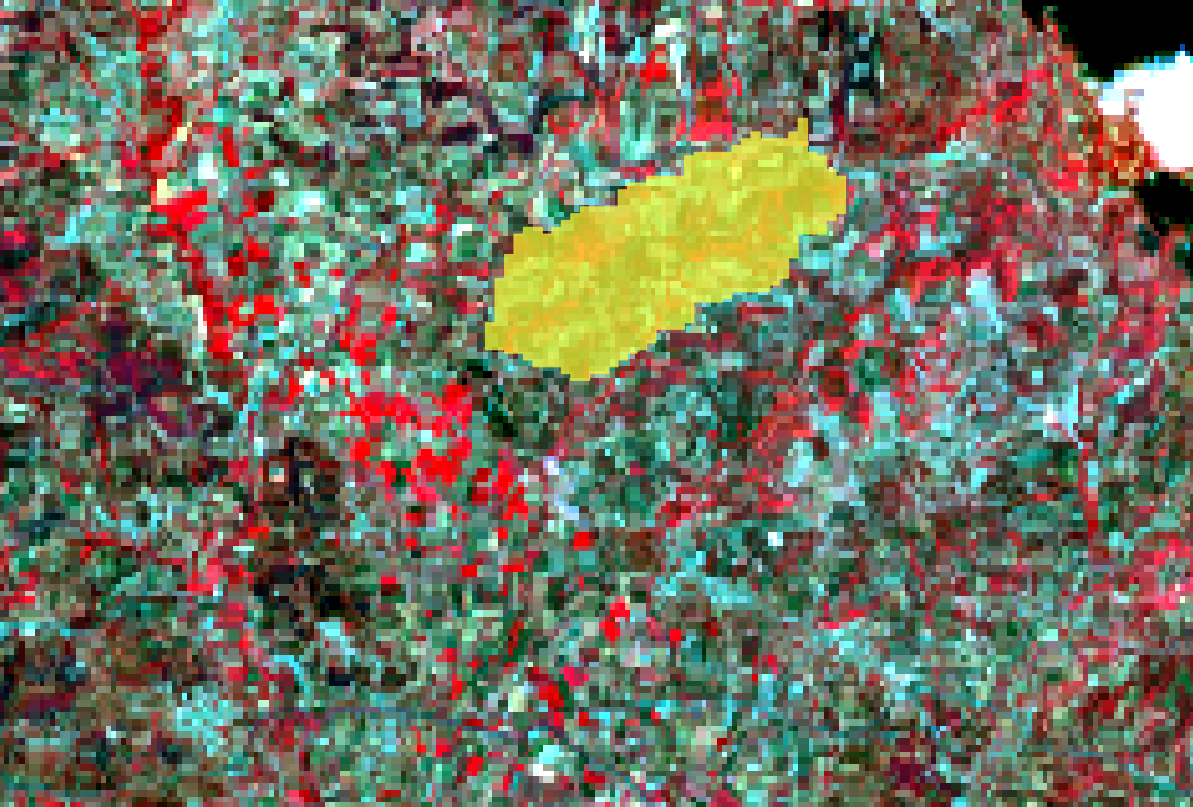}%
\label{res:829_pred}}
\caption{Post-fire Sentinel-2 caption for a wildfire in Heraklia, Peloponnese, Greece, along with (a) MODIS post-event image (8/8/2025), (b) CEMS delineation (9/8/2025), (c) BAM-MRCD delineation. (b) and (c) include the corresponding Sentinel-2 image as background. BAM-MRCD utilized a MODIS post-fire caption on 8/8/2025, whereas the first available Sentinel-2 caption was on 9/8/2025. The wildfire affected approximately 622.7 ha. We observe that the model was not confused by the agricultural land which surrounds the burnt area.}
\label{res:829}
\end{figure*}

\begin{figure*}[!t]
\captionsetup[subfloat]{captionskip=10pt}
\centering
\subfloat[]{\includegraphics[width=2.2in]{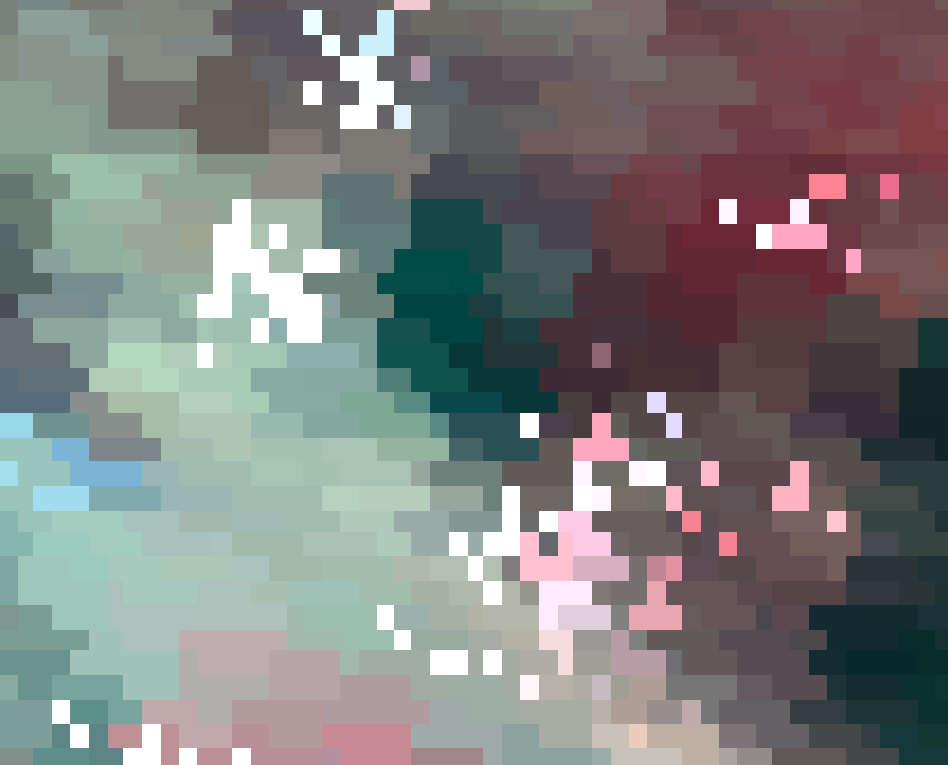}%
\label{res:825_mod}}
\hspace{3.5mm}%
\subfloat[]{\includegraphics[width=2.2in]{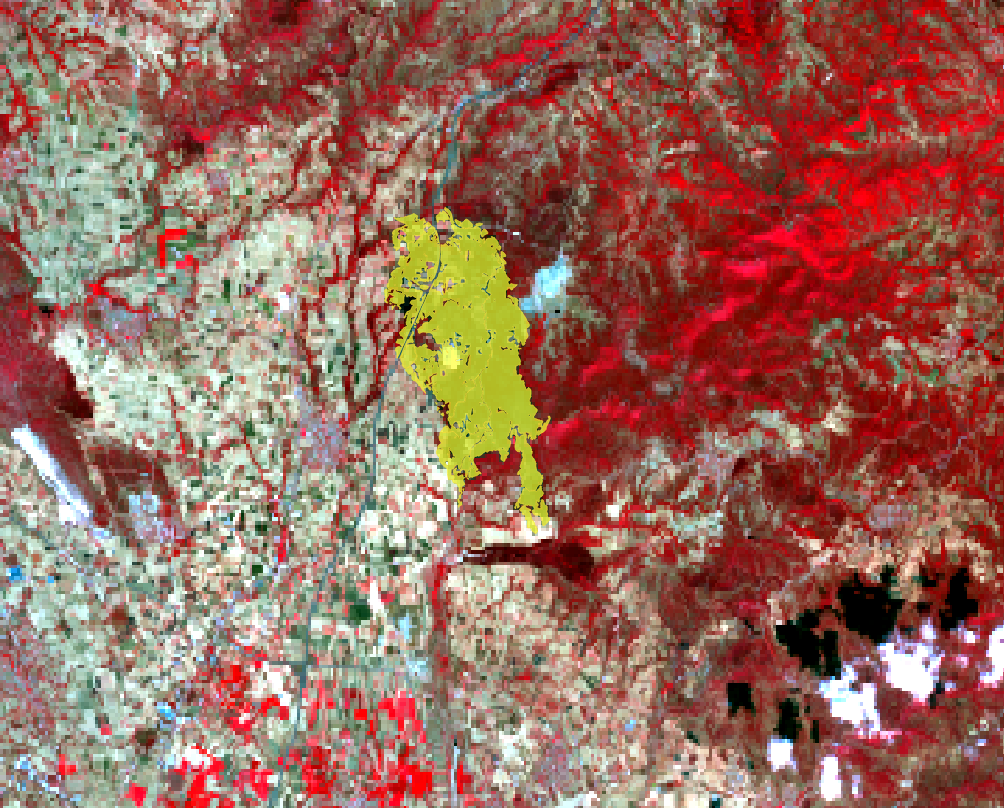}%
\label{res:825_cems}}
\hspace{3.5mm}%
\subfloat[]{\includegraphics[width=2.2in]{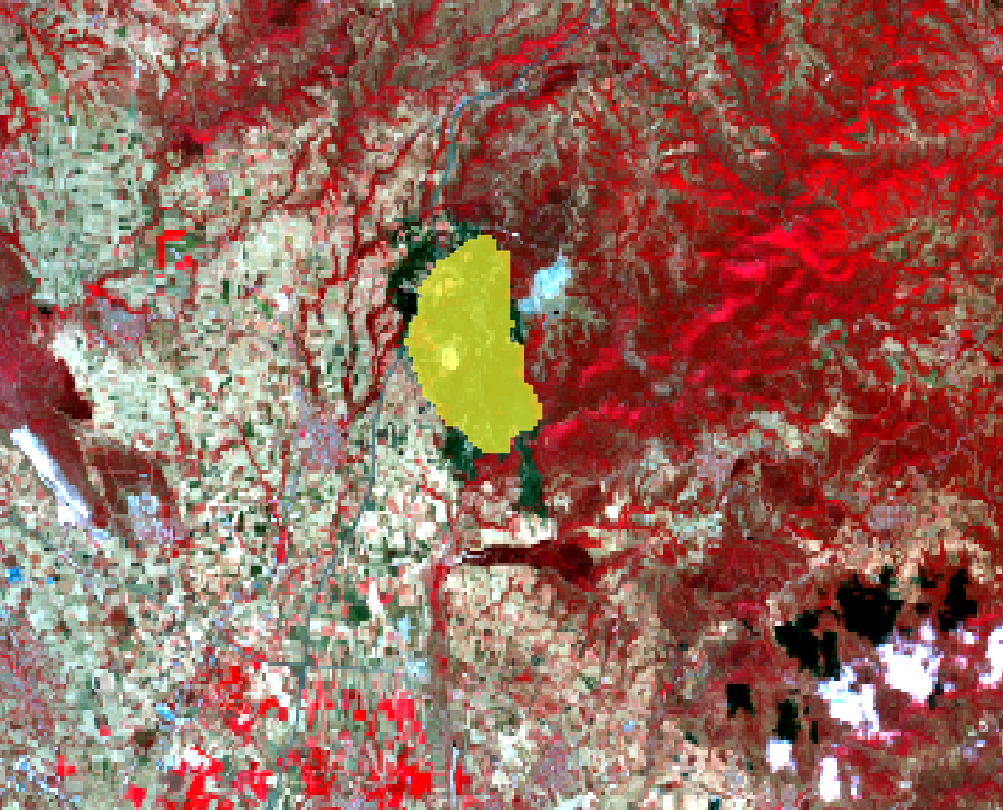}%
\label{res:825_pred}}
\caption{Post-fire Sentinel-2 caption for a wildfire in Assiros, Central Macedonia, Greece, along with (a) MODIS post-event image (3/8/2025), (b) CEMS delineation (2/8/2025), (c) BAM-MRCD delineation. (b) and (c) include the corresponding Sentinel-2 image as background. BAM-MRCD utilized a MODIS post-fire caption on 3/8/2025, whereas the first available Sentinel-2 caption was on 7/8/2025. The wildfire affected approximately 831.9 ha. BAM-MRCD managed to capture the main outline of the burn scar despite its oblong shape and limited information across the horizontal axis.}
\label{res:825}
\end{figure*}

\begin{figure*}[!t]
\captionsetup[subfloat]{captionskip=10pt}
\centering
\subfloat[]{\includegraphics[width=2.2in]{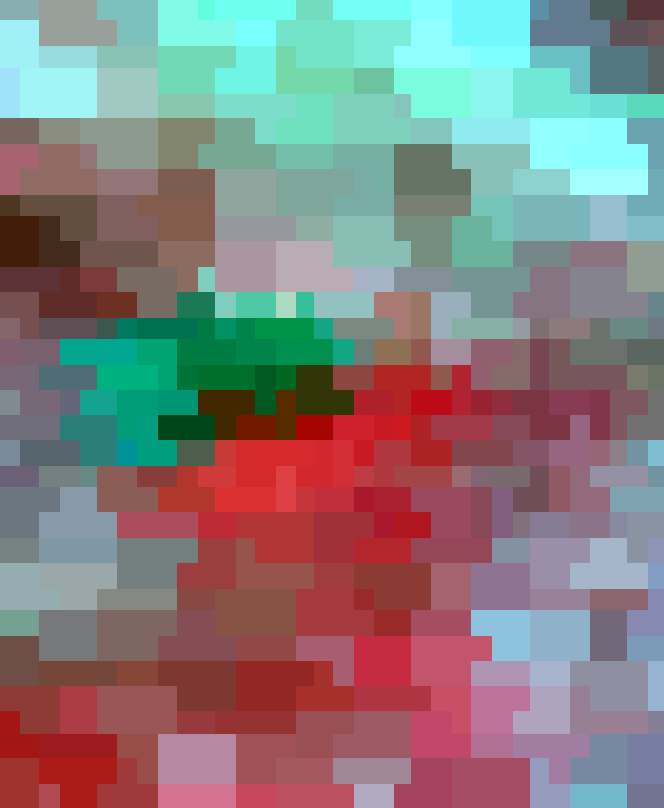}%
\label{res:821_mod}}
\hspace{3.5mm}%
\subfloat[]{\includegraphics[width=2.2in]{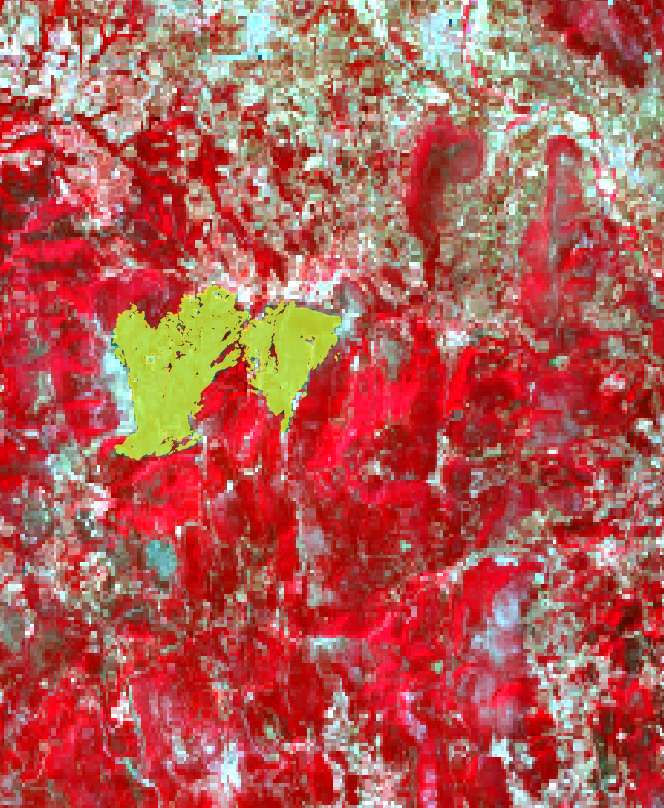}%
\label{res:821_cems}}
\hspace{3.5mm}%
\subfloat[]{\includegraphics[width=2.2in]{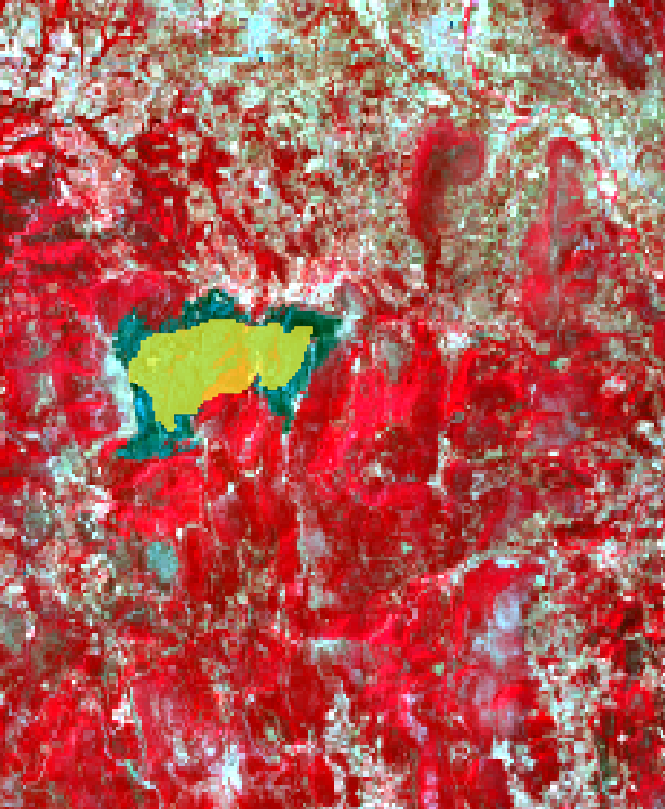}%
\label{res:821_pred}}
\caption{Post-fire Sentinel-2 caption for a wildfire in Aetos, Peloponnese, Greece, along with (a) MODIS post-event image (27/7/2025), (b) CEMS delineation (28/7/2025), (c) BAM-MRCD delineation. (b) and (c) include the corresponding Sentinel-2 image as background. BAM-MRCD utilized a MODIS post-fire caption on 27/7/2025, whereas the first available Sentinel-2 caption was on 28/7/2025. The wildfire affected approximately 650.2 ha. BAM-MRCD captured the main body of the burn scar, missing the narrower vertices that corresponded to very few MODIS pixels.}
\label{res:821}
\end{figure*}

In Fig. \ref{res:832}-\ref{res:821} we present a number of recent wildfires in Greece that took place over the 2025 fire season and were not included in the FLOGA dataset. These wildfires have been mapped by the Copernicus Emergency Management Service (CEMS) and the corresponding delineations are used as ground truth. Each example depicts the post-fire Sentinel-2 image closest to the wildfire (in a NIR-Red-Green composite), with the CEMS delineation vis-à-vis the prediction of BAM-MRCD. We also provide the post-fire MODIS caption that was used as input in our model. Fig. \ref{res:832} displays the capability of BAM-MRCD to correctly retrieve the perimeter of a wildfire in case of a rather large burn scar, where multiple MODIS pixels provide useful information on the impacted area. Fig. \ref{res:829} presents the case of a smaller burn scar in a highly complex terrain where the fire mainly affected agricultural land. Our proposed model has successfully managed to locate the event and provide a generic outline of the affected region, omitting a small area in the north comprising weakly connected burnt parcels. Moreover, the model proved its robustness against land covers with spectral signatures similar to charred land, such as agricultural parcels in this particular case, which were ignored and misclassifications were avoided.

One limitation observed in the performance of the proposed model pertains to the accurate segmentation of certain burn scars that exhibit narrow or oblong shapes. Such examples can be seen in Fig. \ref{res:825} and \ref{res:821}; the former has a rather narrow oblong shape, whereas the latter comprises two tangent triangular shapes. A potential explanation for this shortcoming lies in the spatial resolution mismatch between the input MODIS imagery (500m GSD) and the desired output at Sentinel-2 spatial resolution (60m GSD). Given that these narrow parts are often only represented by a small number of MODIS pixels across their diameter, the sensor may fail to capture sufficient detail or spatial context within these regions. Furthermore, the coarse spatial resolution of MODIS likely leads to significant spatial aggregation, which may blur the boundaries of these narrow or elongated burn scars. As a result, these areas may not provide the full range of spectral information necessary for the model to distinguish them accurately, especially when the features are sub-pixel in size. Additionally, the delineation in Fig. \ref{res:825} may be further constrained by the presence of clouds which obstruct the southern part of the burn scar. Finally, we must also note that in all the aforementioned cases BAM-MRCD was able to provide a quick estimation of the impacted area one day before the next available cloud-free Sentinel-2 caption, with the exception of Fig. \ref{res:825} where the utilized MODIS caption was 5 days in advance.

\begin{figure*}[!t]
\captionsetup[subfloat]{captionskip=10pt}
\centering
\subfloat[]{\includegraphics[width=2.2in]{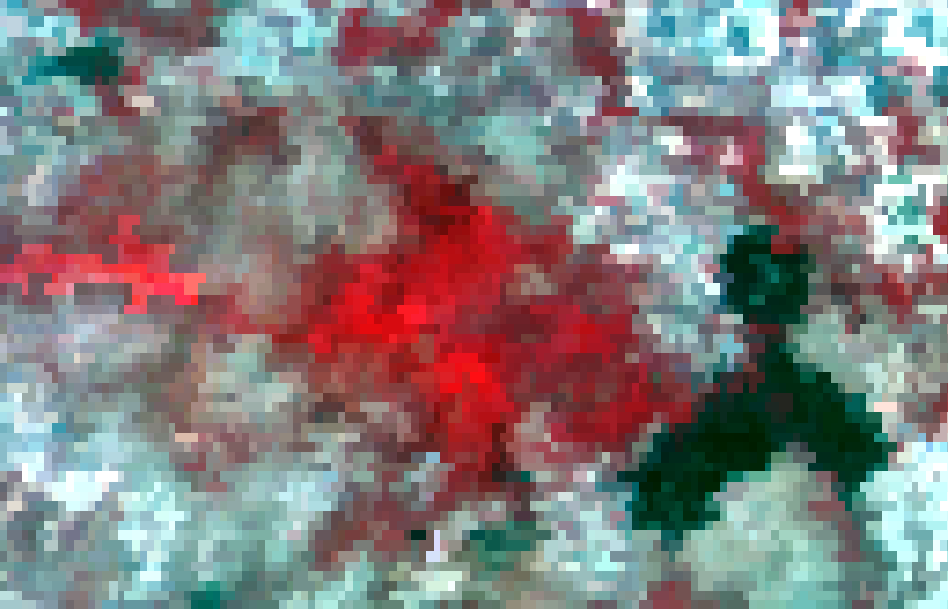}%
\label{res:822_mod}}
\hspace{3.5mm}%
\subfloat[]{\includegraphics[width=2.2in]{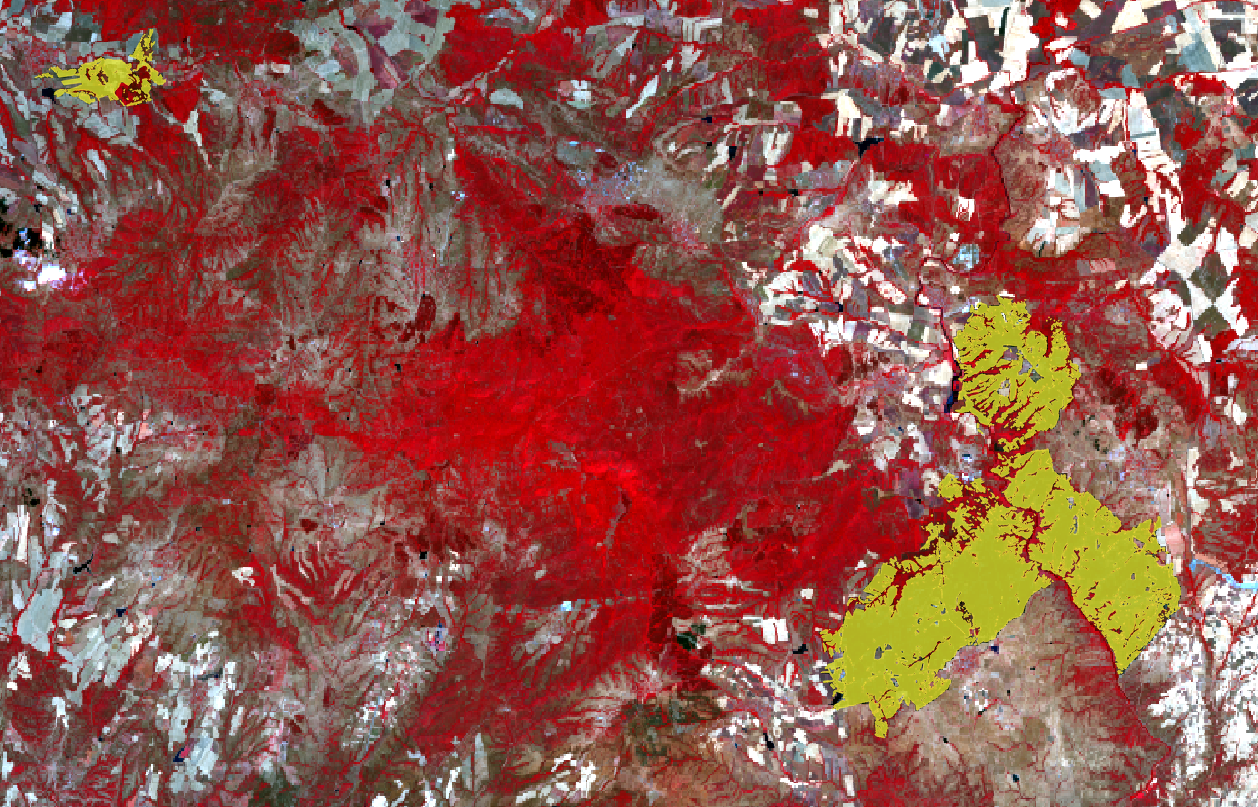}%
\label{res:822_cems}}
\hspace{3.5mm}%
\subfloat[]{\includegraphics[width=2.2in]{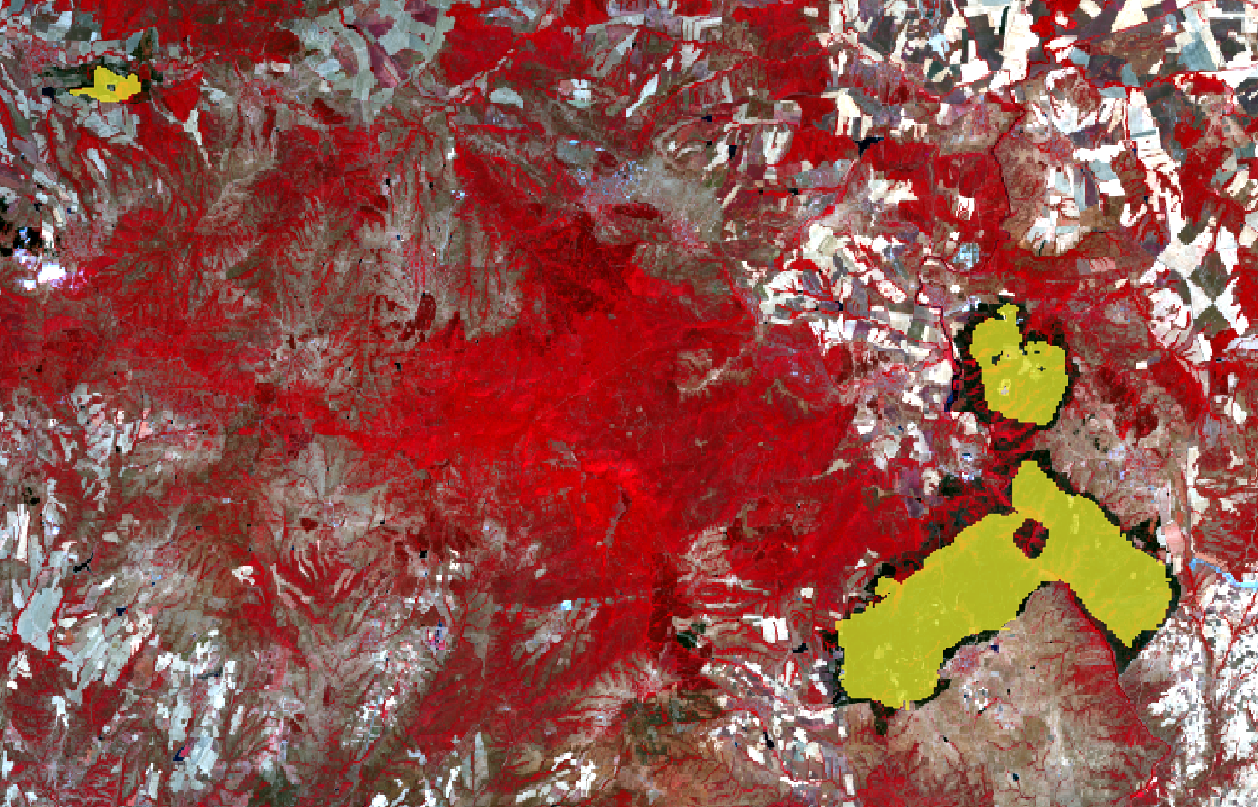}%
\label{res:822_pred}}
\caption{Post-fire Sentinel-2 caption for two simultaneous wildfires in Lesovo, Bulgaria, along with (a) MODIS post-event image (30/7/2025), (b) CEMS delineation (30/7/2025), (c) BAM-MRCD delineation. (b) and (c) include the corresponding Sentinel-2 image as background. BAM-MRCD utilized a MODIS post-fire caption on 30/7/2025, whereas the first available Sentinel-2 caption was on 4/8/2025. The wildfire affected approximately 6303.4 ha. The model has successfully captured both burnt areas.}
\label{res:822}
\end{figure*}

\begin{figure*}[!t]
\captionsetup[subfloat]{captionskip=10pt}
\centering
\subfloat[]{\includegraphics[width=2.2in]{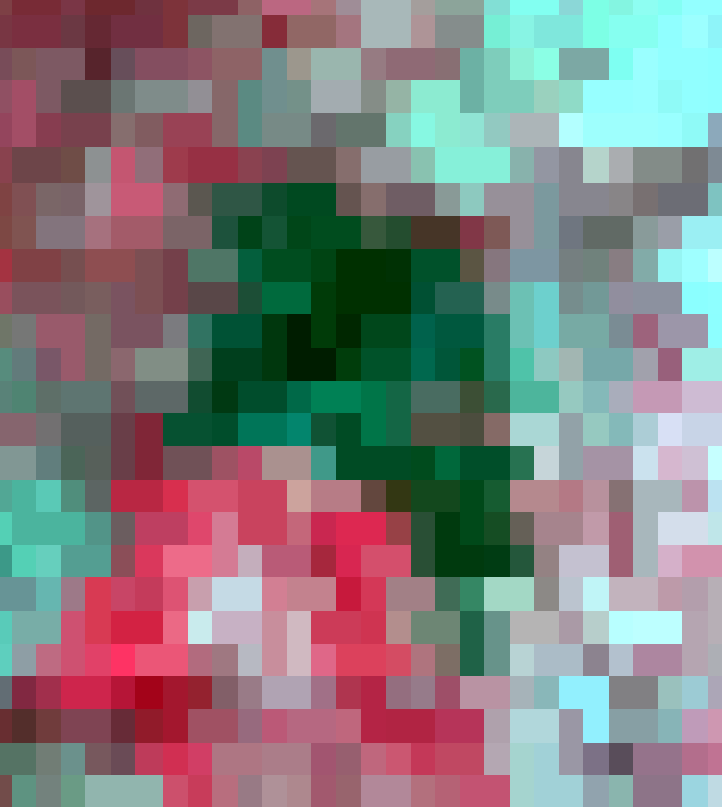}%
\label{res:824_mod}}
\hspace{3.5mm}%
\subfloat[]{\includegraphics[width=2.2in]{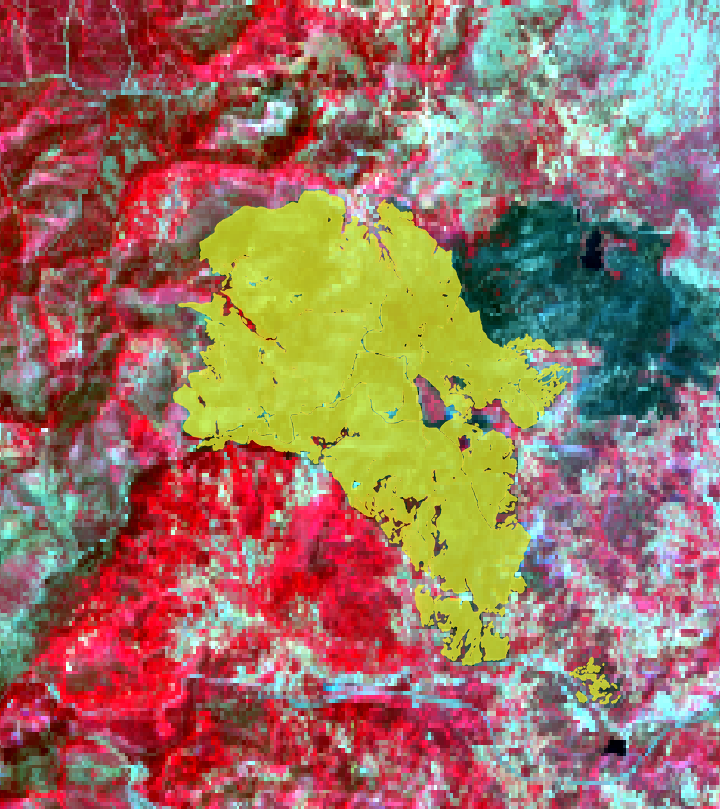}%
\label{res:824_cems}}
\hspace{3.5mm}%
\subfloat[]{\includegraphics[width=2.2in]{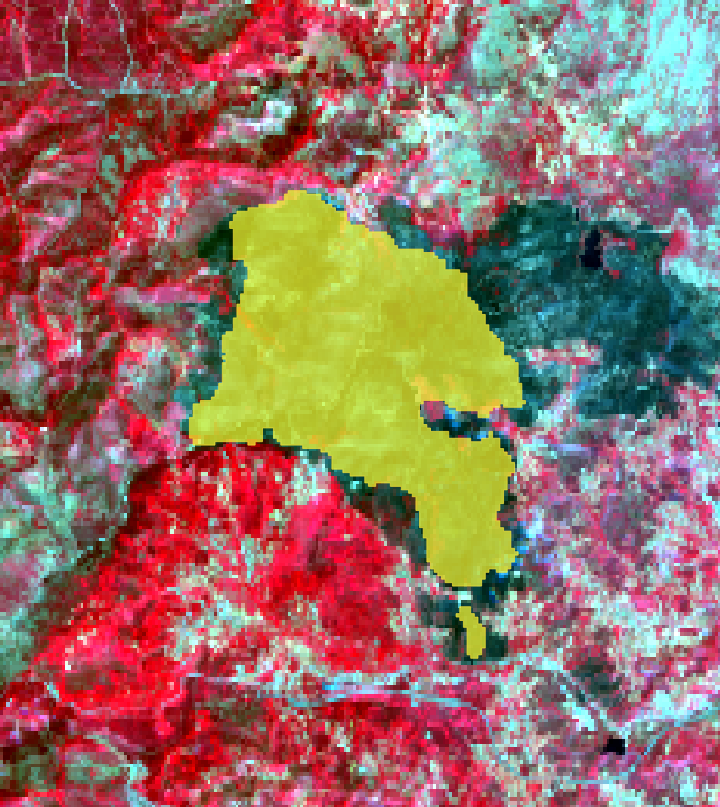}%
\label{res:824_pred}}
\caption{Post-fire Sentinel-2 caption for a wildfire in Vila Real, Portugal, along with (a) MODIS post-event image (7/8/2025), (b) CEMS delineation (7/8/2025), (c) BAM-MRCD delineation. (b) and (c) include the corresponding Sentinel-2 image as background. BAM-MRCD utilized a MODIS post-fire caption on 7/8/2025, whereas due to malfunction on 3/8/2025 the first usable Sentinel-2 caption was on 13/8/2025. The wildfire affected approximately 2891.9 ha. We also observe an older burn scar in the middle right region of the image, which was correctly ignored by the model since it was visible in both the pre- and post-fire input imagery.}
\label{res:824}
\end{figure*}

To explore our model's ability to generalize in areas outside of Greece, we provide examples of wildfire events in Bulgaria and Portugal during the same 2025 fire season as mapped by CEMS. Fig. \ref{res:822} displays the aftermath of two concurrent wildfires in eastern Bulgaria. BAM-MRCD correctly detected both events and provided a rough estimation of the corresponding polygons 5 days before the first available Sentinel-2 caption over this area. Furthermore, Fig. \ref{res:824} shows a large wildfire in Portugal which affected a region close to an older burn. BAM-MRCD correctly isolated the most recent burn scar which manifested as an abrupt change between the bitemporal input imagery. Moreover, in this particular case the Sentinel-2 sensor malfunctioned and the next caption on 3/8/2025 was severely corrupted and thus rendered unusable for this task. This resulted in a further delay until the next available caption on 13/8/2025. However, by exploiting the high temporal frequency of the MODIS satellites, BAM-MRCD is able to provide a burn scar map 6 days in advance.

\section{Discussion}
\label{sec:discussion}

Driven by climate change, wildfires have been steadily increasing in frequency and intensity over the previous years, and particularly in the wider Mediterranean basin where drier and hotter weather conditions and elongated summers are projected to become the new norm. As a direct result, Greece has already displayed intense fire seasons with numerous devastating wildfires impacting its territories the past two decades, with the most notable event being the Evros wildfire in August 2023 which was estimated to be the largest in the European Union so far, claiming $\sim$92,000 ha (920 $km^2$) and several lives.

Satellite data have been an invaluable source of information for assessing the impact of natural disasters, since they provide continuous coverage over vast areas, and can offer information even for inaccessible regions. In addition, the recent advancements in deep learning techniques have propelled innovative solutions in several computer vision tasks, often managing to surpass human performance. When transferred to the field of remote sensing, adapting to more complex data regimes, these techniques achieve impressive results. For example, BAM-CD, a deep learning model proposed for burn scar mapping with Sentinel-2 bitemporal imagery, is currently the state-of-the-art technique for this task, surpassing well established operational burn scar mapping methods and other machine/deep learning approaches.

In this study, we adapt BAM-CD to multi-resolution, multi-source input for a rapid, next-day delineation of the impacted area. Taking advantage of the trade-off between spatial and temporal resolution which characterizes satellite sensors, our proposed model, BAM-MRCD, employs a Sentinel-2 pre-fire image to retrieve high detail information of the underlying scene, and bitemporal MODIS imagery to assist the change detection task. By adding an auxiliary target and a deep supervision loss, the model manages to accurately delineate burn scars of varying size, as well as retrieve a significant number of wildfire events from the test set. Our experiments on a great number of unseen events and regions showcase the superior performance of BAM-MRCD when compared with other CD and SR-CD approaches, whereas predictions on more recent events that were not included in the test set highlight the model's potential to be successfully adopted in operational settings where a rapid assessment of the impacted area is of vital importance.

This formulation of BAM as a multi-resolution, multi-source change detection problem with unique characteristics is another key contribution of this work. Distinct from other change detection tasks in computer vision, burn scar mapping aims for the delineation of non homogeneous changes that lack strict borders and geometry, and are expressed in different ranges of the e/m spectrum. Furthermore, the particular problem discussed in this work employs bitemporal imagery from two different satellites, with vastly disparate spatial, temporal and spectral resolutions. All these challenges render BAM SR-CD a rather complex task, requiring careful handling of the data and meticulous design of the model architecture. During the development process, it also became evident that established segmentation metrics, such as IoU, cannot adequately assess the model's performance on changes of smaller scale. Therefore, we proposed a multi-scale version of IoU which better captures the strengths and weaknesses of the examined approaches, and offers important insights on the ability of a model to map events of varying size.

However, we acknowledge a number of weaknesses in the proposed approach. First, despite the impressive results achieved by BAM-MRCD, there is still room for improvement, particularly in the delineation of small- and medium-sized events, or burn scars with small diameter across at least one dimension. In addition, the exclusive use of multispectral satellite imagery imposes a significant restriction on the usability of the method during cloud and/or smoke coverage. Such adverse atmospheric conditions could delay the employment of the model until the next available cloud-free MODIS image, since clear pre-fire imagery can be retrieved at previous points in time.
Finally, the MODIS satellite constellation is planned to be retired late 2026/early 2027, with the Visible Infrared Imaging Radiometer Suite (VIIRS) satellites taking their place \cite{wong_modis_2024}. However, we expect that our approach will still be applicable since there is a rich archive of VIIRS data to train from, and both constellations share the same basic characteristics.

\section{Conclusion}
\label{sec:conclusion}

In this work, we formulated burn scar mapping as a distinct and rather challenging task, which requires careful handling of data and architectural design. We showcased that next-day delineation of wildfire impacted areas can be achieved by exploiting the complementary strengths of different satellite sources. In particular, we examined the synergy of Sentinel-2 and MODIS imagery for a timely mapping at 60m GSD. To that end, an extension of the state-of-the-art BAM-CD model was proposed, namely BAM-MRCD, which displays superior performance against solid baselines and change detection approaches. A number of more recent unseen events from Greece, Bulgaria and Portugal were examined and the results demonstrated the generalization capabilities of BAM-MRCD as well as its robustness against diverse landscapes. Finally, difficult cases of reduced model performance were discussed, and the general limitations of our work were highlighted. We are confident that the proposed approach is a promising step towards the rapid production of burn scars during emergency situations and we hope it will pave the road for further research in this emerging field.

\bibliographystyle{IEEEtran}
\bibliography{references}

\end{document}